\documentclass[sigconf,nonacm]{aamas}

\usepackage{balance}
\usepackage{fvextra}
\usepackage{needspace}
\setcopyright{none}
\AtBeginDocument{%
  \fancypagestyle{standardpagestyle}{\fancyhf{}\fancyfoot[C]{\footnotesize\thepage}}%
  \fancypagestyle{firstpagestyle}{\fancyhf{}\fancyfoot[C]{\footnotesize\thepage}}%
  \pagestyle{standardpagestyle}%
}

\title[Interchangeable in Role, Not in History]{Interchangeable in Role, Not in History: Replacing Members of Formed LLM Teams}

\author{Jianxin Gao}
\email{jxgao@cau.edu.cn}
\affiliation{\institution{China Agricultural University}}
\author{Runze Li}
\email{rzli5524@mails.jlu.edu.cn}
\affiliation{\institution{Jilin University}}
\author{Tianyi Yu}
\email{tiantyu318@stu.tjufe.edu.cn}
\affiliation{\institution{Tianjin University of Finance and Economics}}
\author{Liangwei Ren}
\email{lwren@cau.edu.cn}
\affiliation{\institution{China Agricultural University}}
\author{Bohan Chen}
\email{chenbohan@cau.edu.cn}
\affiliation{\institution{China Agricultural University}}
\author{Zining Wang}
\email{25101314@mail.tust.edu.cn}
\affiliation{\institution{Tianjin University of Science and Technology}}
\renewcommand{\shortauthors}{Jianxin Gao et al.}
\makeatletter
\def\@mkauthors@iii{%
  \global\setbox\mktitle@bx=\vbox{\unvbox\mktitle@bx
    \centering\rmfamily
    {\fontsize{12}{14.5}\selectfont
    Jianxin Gao\textsuperscript{1,*,$\dagger$}\quad
    Runze Li\textsuperscript{2,*}\quad
    Tianyi Yu\textsuperscript{3,*}\quad
    Liangwei Ren\textsuperscript{1}\quad
    Bohan Chen\textsuperscript{1}\quad
    Zining Wang\textsuperscript{4}\par}
    \vspace{4pt}
    {\fontsize{10}{12.5}\selectfont
    \textsuperscript{1}China Agricultural University\quad
    \textsuperscript{2}Jilin University\par
    \textsuperscript{3}Tianjin University of Finance and Economics\quad
    \textsuperscript{4}Tianjin University of Science and Technology\par}
    \vspace{4pt}
    {\fontsize{9}{11}\selectfont
    \href{mailto:jxgao@cau.edu.cn}{jxgao@cau.edu.cn}\quad
    \href{mailto:rzli5524@mails.jlu.edu.cn}{rzli5524@mails.jlu.edu.cn}\quad
    \href{mailto:tiantyu318@stu.tjufe.edu.cn}{tiantyu318@stu.tjufe.edu.cn}\par
    \href{mailto:lwren@cau.edu.cn}{lwren@cau.edu.cn}\quad
    \href{mailto:chenbohan@cau.edu.cn}{chenbohan@cau.edu.cn}\quad
    \href{mailto:25101314@mail.tust.edu.cn}{25101314@mail.tust.edu.cn}\par
    \vspace{2pt}
    \textsuperscript{*}Equal contribution.\quad
    \textsuperscript{$\dagger$}Corresponding author: Jianxin Gao.\par}
    \vspace{20.21145pt}}
}
\makeatother

\begin{abstract}
Multiagent LLM systems often treat a role as a slot: when a member is replaced, any agent competent at the role is expected to fill it. We test this view on teams whose members keep short written notebooks. A one-seat swap replays a formed, frozen team with its own member, with a same-role newcomer carrying another team's notebook for that role, or with a newcomer carrying none. In cooperative cooking, wildfire response and medical question answering, the experienced newcomer, averaged over seats, outscores the one without a notebook in every setting, so the usual comparison would approve it. Against the original team, the same replacement falls short in every setting, and the team spends more messages per unit of progress. On the hardest cooking level, the shortfall appears only after the team has worked together, persists across model families and decoding temperatures, and narrows by two thirds with added planning-only members. Because the team's history is written down, we can take it apart. In the default configuration on that level, notebooks from the same donor teams leave a newcomer that leads, holding the recipe, worse off than no notebook and one that follows better off. The notebook lines about the partner reproduce this whole difference between the seats. Whether an agent can replace a teammate depends on the seat and the team it enters.
\end{abstract}


\newcommand{\CI}[2]{[#1, #2]}
\newcommand{\Dseat}{\Delta_{\mathrm{seat}}}
\newcommand{\bO}{\mathrm{O}}
\newcommand{\bE}{\mathrm{E}}
\newcommand{\bF}{\mathrm{F}}

\begin{document}
\maketitle
\hypersetup{pdfauthor={Jianxin Gao, Runze Li, Tianyi Yu, Liangwei Ren, Bohan Chen, Zining Wang}}
\section{Introduction}\label{sec:intro}
Multiagent systems built from large language models (LLMs) often replace members: a model version is retired, a process fails, or a better specialist becomes available. The frameworks behind these systems make replacement easy. A role is specified like a box on an organization chart, with a responsibility, an interface and a place in a communication topology, as in organizational models of multiagent systems \cite{ferber1998metamodel,hubner2002moise,zambonelli2003gaia}, and the agent occupying it is addressed as an instance of the role rather than as an individual \cite{hong2024metagpt,qian2024chatdev,wu2024autogen}. Methods that optimize a team go further and recruit, score or drop agents as units \cite{chen2024agentverse,liu2024dylan,wang2025agentdropout}. These designs assume that any agent competent at a role can fill it, and so imply that a formed team is indifferent to which one does.

\begin{figure*}
\includegraphics[width=\textwidth]{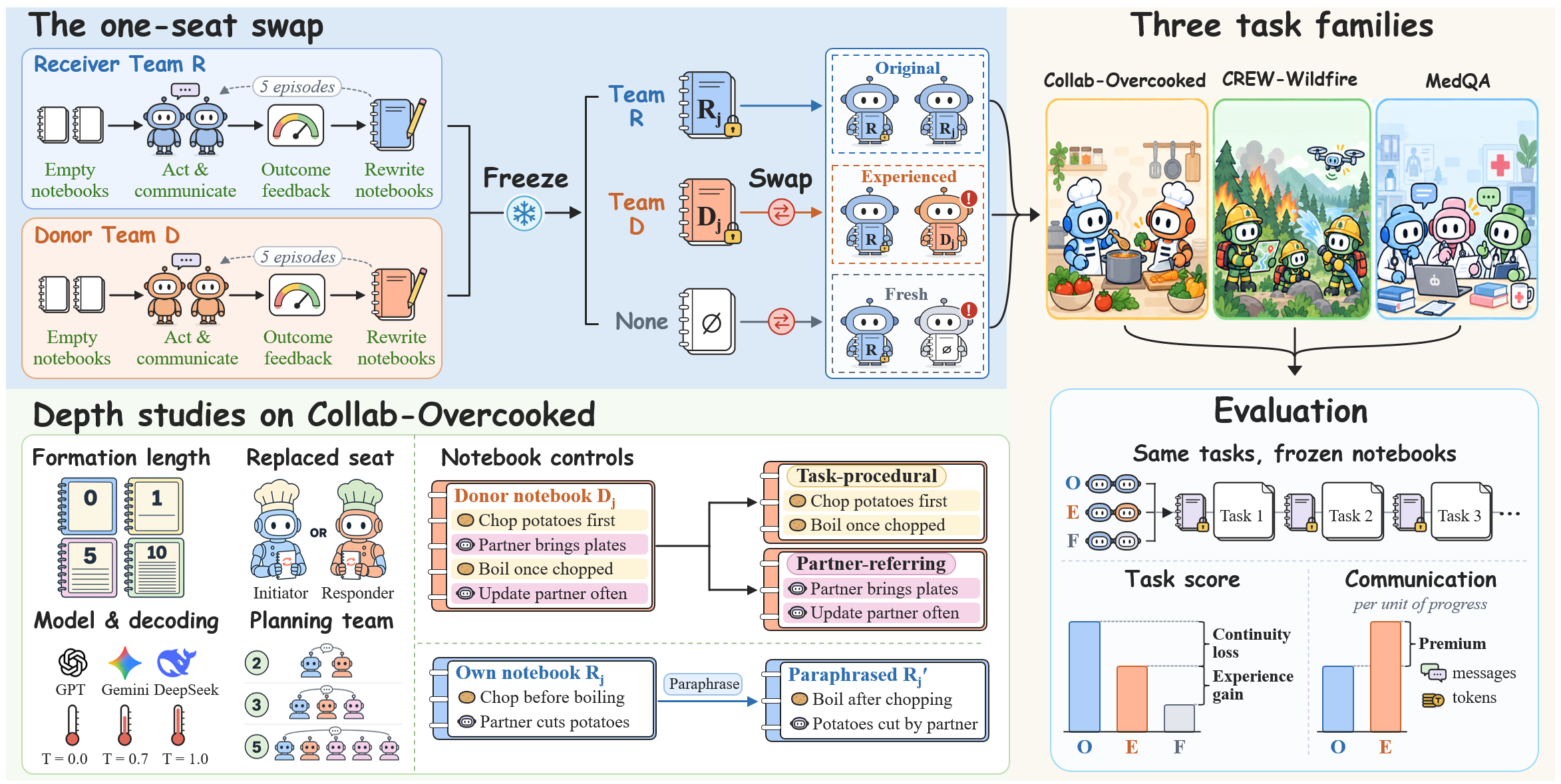}
\caption{The one-seat swap. Receiver R and donor D form separately and are frozen. R's seat $j$ is then replayed with its own notebook $R_j$ (Original, O), with D's notebook $D_j$ (Experienced, E) or with none (Fresh, F). Red markers show the notice. Bars and notebook lines are illustrative.}
\Description{Schematic of the one-seat swap protocol comparing Original, Experienced and Fresh branches after receiver and donor teams form independently.}
\label{fig:swap}
\end{figure*}

In LLM teams, the experience of working together can be written down. Reflection and memory mechanisms let an agent carry what it learned from one episode to the next without touching its weights \cite{park2023generative,shinn2023reflexion,xu2025amem}, and a growing line of work pools such memory across trials or shares it between agents \cite{chang2026memcollab,zhang2025gmemory}. What accumulates is more than task procedure. A memory can also record who tends to move first, which requests get answered and when information arrives unasked. After five cooking episodes in our study, one agent noted that in the pumpkin episode ``re-staging slices at [2,2] and dish at [2,3] enabled successful handoff.'' The note records a handoff arrangement that worked with one partner, and a member of another team never agreed to it. Research on human teams treats such knowledge of one's teammates as part of transactive memory and finds that its benefits can turn into detriments when membership changes \cite{lewis2007group,wegner1987transactive}. Replacing a member with an experienced outsider therefore trades the receiving team's history of working together for experience that the newcomer earned elsewhere \cite{huckman2009familiarity,kane2005rotation,reagans2005experience}.

Treating a role as a slot makes two predictions. A newcomer that matches the incumbent in role, model and amount of experience should leave the team as strong as before. And a memory should carry a fixed value, helping or hurting by the same amount in whichever seat and team it is installed. The comparison that evaluations typically make tests neither prediction. It mirrors how memory mechanisms are usually judged, by removing the memory \cite{park2023generative,shinn2023reflexion,xu2025amem}. Setting an experienced newcomer against one without memory asks whether the newcomer's history is worth anything. It cannot ask what the receiving team lost, because the team as it was is absent from the comparison. A task score reported alone hides a further cost, since a replacement can buy back score by asking more questions, sending longer messages or drawing retained teammates into more exchange.

\paragraph{A one-seat swap} We test both predictions with one intervention (Figure~\ref{fig:swap}). Two teams form independently in the same task setting, playing the same number of formation episodes, and are then frozen. Each agent's memory is a short private notebook, and we call one role's position in a particular team a \emph{seat}. From the receiving team's saved state we replay three branches that differ in one seat only: \emph{Original} keeps the incumbent, \emph{Experienced} installs a same-role newcomer carrying the notebook written by the donor team's member in that role, and \emph{Fresh} installs a same-role newcomer with an empty notebook. Retained teammates start every branch from byte-identical notebooks, and every branch faces the same task list. Two quantities then separate: the \emph{continuity loss} $\bO-\bE$, what the team still loses with an experienced replacement, and the \emph{experience gain} $\bE-\bF$, what the imported notebook is worth. We read both beside the communication the team spends per unit of progress.

Replacement poses a different problem from ad hoc teamwork, which asks how an agent can cooperate with partners it has never met \cite{mirsky2022survey,stone2010adhoc}. Here the partners have met each other, and only the newcomer has not met them. Original and Experienced correspond to the self-play and cross-play conditions of zero-shot coordination \cite{hu2020otherplay,strouse2021collaborating}, which compare partners that trained together with partners that trained apart. Two differences make the swap more than a cross-play test. The history is a short document rather than a set of weights, so it can be emptied, reworded, split and transplanted one part at a time, and the swap can locate where in the history a cost lies. And a replacement is asymmetric: the newcomer is told that it is joining a team, while its teammates carry on from their own history.

\paragraph{Contributions} The paper makes three contributions.
\begin{itemize}
\item \emph{A protocol for evaluating replacement after formation.} The one-seat swap separates the continuity loss from the experience gain, reports communication per unit of progress beside the task score, and can be added to any benchmark that saves agent state (Section~\ref{sec:swap}).
\item \emph{A cost that the usual comparison misses.} In five settings from three task families, the experienced newcomer, averaged over seats, beats a fresh one yet falls short of the member it replaces, and the team needs 5\% to 34\% more messages per unit of progress (Section~\ref{sec:screen}). On L3, the hardest Collab-Overcooked level, a team swapped before formation shows no detectable loss. After ten formation episodes, about half of what formation built leaves with the replaced member, and the imported notebook recovers about an eighth, the only part that a comparison with an empty notebook can see. Changing the model family or the decoding temperature leaves the loss in place (Section~\ref{sec:formation}).
\item \emph{An account of where the cost lies.} On L3, in the default configuration, notebooks from the same donor teams leave a newcomer initiator, who holds the recipe, 7.98 points below a fresh newcomer and a newcomer responder 7.05 points above one, a seat asymmetry with the same sign in all ten repetitions. The notebook lines that refer to the partner reproduce this asymmetry in full, the lines about the task reproduce none of it, and rewording an incumbent's own notebook costs almost nothing (Sections~\ref{sec:seat} and~\ref{sec:content}). Adding planning-only members cuts the loss from 9.81 to 3.41 points (Section~\ref{sec:planning}).
\end{itemize}
For system builders, the results call for evaluating a replacement in its receiving team and per seat, for transferring task procedure, which did not hurt in either seat, and for checking what a notebook says about partners against the seat it will enter.

\section{Related Work}
\paragraph{Evaluating LLM teams} AgentBench scores LLMs as interactive agents \cite{liu2024agentbench}, and LLM-Coordination and MultiAgentBench probe coordination, collaboration and competition directly \cite{agashe2025llmcoordination,zhu2025multiagentbench}. Collab-Overcooked enforces asymmetric information and interdependent actions \cite{sun2025collabovercooked}, and CREW-Wildfire and MedAgentBoard add spatial teamwork and multiagent medical question answering \cite{hyun2025crewwildfire,zhu2025medagentboard}. Failure taxonomies separate specification, inter-agent alignment and task execution \cite{cemri2025fail}, and adaptive systems can reach acceptable final accuracy while the interaction between their agents breaks down \cite{so2026superficial}. Resilience studies make one agent faulty and compare how hierarchical and flat structures absorb the damage \cite{huang2025resilience}. In all of these settings, the agents that begin an evaluation finish it. We use three of the benchmarks and vary membership instead, with a newcomer that is competent and differs in where its experience was earned.

\paragraph{Roles and portable memory} Role-based frameworks distribute the ReAct loop \cite{yao2023react} across specialists, such as writers and reviewers in MetaGPT, ChatDev and AutoGen \cite{hong2024metagpt,qian2024chatdev,wu2024autogen} or a selector, decomposer and refiner in MAC-SQL \cite{wang2025macsql}. The narrow interface between roles is what makes a role-matched replacement a coherent operation. Reflexion, generative agents and A-Mem retain and organize the state an agent accumulates \cite{park2023generative,shinn2023reflexion,xu2025amem} and are evaluated on how much it helps the agent that owns it. Experiential co-learning measures how two roles' accumulated shortcuts affect their product and their talk \cite{qian2024colearning}, G-Memory carries a multiagent system's collaboration history across trials \cite{zhang2025gmemory}, and MeCo caches a multi-robot team's plans for similar tasks \cite{wang2026meco}. Memory that travels can also mislead. Agents follow a retrieved record closely enough that a mismatched one hurts the execution it was meant to help \cite{xiong2026memory}, and a memory written with one model carries that model's reasoning style into another, which MemCollab counters by distilling task-level invariants from contrasting trajectories \cite{chang2026memcollab}. In our swap each notebook stays with the model that wrote it, and the question is whether a notebook that helped in one team still helps in another.

\paragraph{Unfamiliar partners and shared history} Ad hoc teamwork studies collaboration without prior coordination \cite{mirsky2022survey,stone2010adhoc}, and open variants let the roster itself change \cite{rahman2023open}. Other-Play, fictitious co-play and zero-shot coordination target compatibility with partners never met before \cite{hu2020otherplay,strouse2021collaborating,treutlein2021zsc}, including in Overcooked and in human-AI teams \cite{carroll2019utility,shih2021conventions}. LLM agents have been found robust to unseen human-proxy partners in Overcooked-AI \cite{agashe2025llmcoordination}, and partner models can be inferred, chosen among and tested after an abrupt change in the partner's behaviour \cite{ni2026tom,abdurahman2026trait}. A replacement changes the team instead: the newcomer's partner model arrives intact, but the partner it describes is gone. In human-AI teams, an update that makes an AI more accurate can lower team performance by breaking what its human partner has learned to expect \cite{bansal2019updates}. In our teams the expectations belong to AI teammates and are written in their notebooks. Research on human teams and collaborative reference separates what an individual knows from what a group has established together \cite{clark1986referring,huckman2009familiarity,reagans2005experience,wegner1987transactive}. LLM populations also converge on shared conventions \cite{ashery2025conventions}, although multimodal LLM agents succeed in reference games without partner-specific conventions of the human kind \cite{wang2026aligned}.

\section{The One-Seat Swap}\label{sec:swap}
\subsection{Agents, notebooks and the intervention}
An agent is a fixed language model with a role instruction, an action interface and a private notebook. Within an episode it observes, acts and sends messages that the team shares. Between formation episodes it rewrites its notebook from its experience and the outcome feedback. The rewrite instruction asks for a cumulative notebook of concise numbered notes, one main point per note, that keeps earlier lessons unless new evidence contradicts them. It prescribes no number of lines and no mix of content, and a single length budget covers what the agent records about the task and about its partners. The notebook is the only state that can travel between teams, and no model weights are updated. Notebooks stay short: the twenty donor notebooks analysed in Section~\ref{sec:content} run five to ten lines, well within a 1,000-token budget, so each can be read, labelled and split line by line.

The protocol begins with two teams that form independently in the same task setting. Each starts from empty notebooks and plays the same number of formation episodes, after which both teams are frozen and their notebooks saved as a formation snapshot. The team with the higher mean formation score becomes the \emph{receiver}, the team with more to lose, and the other becomes the \emph{donor}, with ties broken by a fixed index. Writing $m^R_j$ and $m^D_j$ for their notebooks at a designated role $j$ ($R_j$ and $D_j$ in Figure~\ref{fig:swap}), and $m^R_{-j}$ for the notebooks of the receiver's other members, the three branches are
\begin{equation}
\mathcal{M}_{\mathrm{O}} = (m^R_{-j}, m^R_j), \quad \mathcal{M}_{\mathrm{E}} = (m^R_{-j}, m^D_j), \quad \mathcal{M}_{\mathrm{F}} = (m^R_{-j}, \varnothing).\label{eq:branches}
\end{equation}
Only the replaced seat changes. Retained teammates enter all three branches with identical notebooks, and outside the model-family comparison the newcomer runs the incumbent's model. What the intervention changes is the \emph{provenance} of one member's experience, while its responsibility, interface and model stay fixed. In this it follows matched designs such as MirrorCraft, whose paired worlds differ only in hidden rules \cite{gao2026mirrorcraft}.

In every evaluation episode the newcomer, experienced or fresh, receives one additional line, \emph{``This is your first episode with this team in the $\langle$role$\rangle$ seat,''} because a notebook written elsewhere would otherwise misinform it about its history with this team. The incumbent and the retained teammates receive no such line, as in a drop-in replacement that leaves the rest of the team untouched. A replacement thus changes two things at once: the history in the seat, and which member knows that the team has changed.

\paragraph{Frozen evaluation} Every branch is evaluated on the same task list, and each evaluation episode reloads that branch's frozen notebooks instead of inheriting updates from the previous episode. Evaluation measures the team at the handover and never adds to its formation history.

\subsection{Two contrasts and the seat asymmetry}
Let $S_c$ be the mean task score of branch $c$ over the evaluation episodes of one formation snapshot. The design yields two signed differences,
\begin{equation}
D_S = S_{\mathrm{O}} - S_{\mathrm{E}}, \qquad G_S = S_{\mathrm{E}} - S_{\mathrm{F}}.\label{eq:contrasts}
\end{equation}
A positive \emph{continuity loss} $D_S$ means the team would have done better keeping its own member. A positive \emph{experience gain} $G_S$ means the imported notebook beat an empty one, and a negative gain means it did harm. With model, role and notice matched, $\bE-\bF$ measures what carrying the donor notebook is worth. $\bO-\bE$ compares the team's own member with this particular replacement and so includes every difference between them, from the notice to the content of the two notebooks, and in the cross-model comparisons of Section~\ref{sec:formation} also the change of model. Together the two contrasts split the drop from Original to Fresh, $\bO-\bF = (\bO-\bE) + (\bE-\bF)$, into the part that the imported notebook recovers and the part it does not.

When one snapshot is branched twice, once replacing the initiator and once the responder, we report the \emph{seat asymmetry}
\begin{equation}
\Dseat = G_S^{\mathrm{resp}} - G_S^{\mathrm{init}} = \bigl(S_{\mathrm{E}}^{\mathrm{resp}} - S_{\mathrm{E}}^{\mathrm{init}}\bigr) + \bigl(S_{\mathrm{F}}^{\mathrm{init}} - S_{\mathrm{F}}^{\mathrm{resp}}\bigr).\label{eq:seat}
\end{equation}
A positive $\Dseat$ means that donor experience is worth more in the responder seat. The two replacements share one Original evaluation, so the first bracket also equals the difference between the two seats' losses, while the second compares the two Fresh branches, which differ in whose own history is missing. $\Dseat$ never uses the Original branch, and the newcomer receives the notice in all four branches it compares, so the notice alone cannot produce it.

Each setting has its own task score from 0 to 100, comparable only within that setting: Progress Completeness (PC) in Collab-Over\-cooked, the share of target trees cut in CREW-Wild\-fire and answer accuracy in MedQA. PC is the benchmark's trajectory-based score and, unlike the progress measure below, charges for redundant interaction \cite{sun2025collabovercooked}.

\subsection{Communication per unit of progress}
Redundant coordination is overhead in one line of work \cite{xie2025acorn} and repair under noise in another \cite{vital2025repair}, so we count message events and token volume separately and scale both by progress. Let $M_{bce}$, $K_{bce}$ and $P_{bce}$ be the shared messages, shared tokens and achieved progress of evaluation episode $e$ in branch $c$ of repetition $b$. Progress is one for a correct final answer and zero otherwise in MedQA, the number of target trees cut in CREW-Wildfire, and $d_{\mathrm{init}} + d_{\mathrm{resp}}$ in Collab-Overcooked, where $d_i$ is the length of the longest prefix of role $i$'s reference actions, as selected by the PC scorer from the benchmark's candidates, that appears as an ordered subsequence of its executed trace. Totals include unsuccessful episodes. We pool within a repetition before dividing,
\begin{equation}
C^{\mathrm{msg}}_{bc} = \frac{\sum_e M_{bce}}{\sum_e P_{bce}}, \qquad C^{\mathrm{tok}}_{bc} = \frac{\sum_e K_{bce}}{\sum_e P_{bce}},
\end{equation}
and average these repetition-level ratios with equal weight to obtain branch-level costs $C_c$. For either measure, the \emph{communication premium} of a replacement is
\begin{equation}
R_C = 100\left(\frac{C_{\mathrm{E}}}{C_{\mathrm{O}}} - 1\right)\%.
\end{equation}
Shared communication excludes private reasoning and notebook updates, a broadcast counts once, and one tokenizer, tiktoken's o200k\_base, counts tokens for all models.

\section{Experiments}
\subsection{Setup}
\paragraph{Task settings} Five settings from three task families carry the study. Collab-Overcooked levels L1 to L3 pair an initiator, who alone holds the recipe and operates the pot, oven and serving, with a responder, who prepares ingredients and dishes and hands them over \cite{sun2025collabovercooked}. The initiator's reference action sequence is the longer of the two on every level, by a ratio of 1.08, 1.13 and 1.46 on L1 to L3. Formation and evaluation stay within one level, whose five recipes are taken in a fixed order with a starting point that rotates across repetitions. Receiver and donor play the same formation sequence and are evaluated on the same five recipes, so the incumbent and the experienced newcomer have played each evaluation recipe equally often. An episode is capped at 1.5$\times$ the reference completion time. CREW-Wildfire puts three acting agents with identical capabilities, one of whom is replaced at random, in a sparse tree-cutting task with six target cells and twelve joint decision rounds \cite{hyun2025crewwildfire}. MedQA~\cite{jin2021medqa} uses the four-agent ColaCare topology of three specialists and a coordinator, which stops at agreement or after three rounds \cite{wang2025colacare,zhu2025medagentboard}. Only specialist seats are replaced, each formation episode is one training question, and evaluation uses test questions.

\begin{table*}
\caption{Breadth screen. Averaged over seats, the experienced newcomer beats a fresh one in every setting yet falls short of the original team. Branch means over ten repetitions. Loss is O $-$ E, gain is E $-$ F, and premiums compare Experienced with Original.}
\label{tab:breadth}
\begin{tabular*}{\textwidth}{@{\extracolsep{\fill}}lrrrrrrrrrrr@{}}
\toprule
& \multicolumn{5}{c}{Task score} & \multicolumn{3}{c}{Messages per unit of progress} & \multicolumn{3}{c}{Tokens per unit of progress} \\
\cmidrule(lr){2-6}\cmidrule(lr){7-9}\cmidrule(l){10-12}
Setting & Orig. & Exp. & Fresh & Loss & Gain & Orig. & Exp. & Prem. (\%) & Orig. & Exp. & Prem. (\%) \\
\midrule
Collab-Overcooked L1 & 93.91 & 92.66 & 89.84 & 1.25 & 2.82 & 2.94 & 3.39 & 15.45 & 46.35 & 53.66 & 15.76 \\
Collab-Overcooked L2 & 84.52 & 81.41 & 77.56 & 3.11 & 3.85 & 5.20 & 6.24 & 19.87 & 88.04 & 104.85 & 19.09 \\
Collab-Overcooked L3 & 71.18 & 64.12 & 57.96 & 7.05 & 6.16 & 8.34 & 11.21 & 34.33 & 163.73 & 218.03 & 33.17 \\
CREW-Wildfire & 88.13 & 85.50 & 83.09 & 2.63 & 2.41 & 3.43 & 3.75 & 9.08 & 82.89 & 89.98 & 8.55 \\
MedQA & 90.71 & 88.42 & 86.46 & 2.29 & 1.96 & 8.83 & 9.29 & 5.31 & 503.10 & 524.02 & 4.16 \\
\bottomrule
\end{tabular*}

\vspace{10pt}
\caption{Configuration study on L3. The default is GPT-5.6 Luna with five formation episodes at $T = 0.7$. Each row replaces the initiator in five repetitions and the responder in the other five, and the five-episode row reuses the seat study's snapshots, taking one seat from each. $X \rightarrow Y$ gives receiving model $X$ and incoming model $Y$. Premiums compare Experienced with Original.}
\label{tab:config}
\begin{tabular*}{\textwidth}{@{\extracolsep{\fill}}llrrrrrrr@{}}
\toprule
& & \multicolumn{5}{c}{Task score (PC)} & \multicolumn{2}{c}{Premium (\%)} \\
\cmidrule(lr){3-7}\cmidrule(l){8-9}
Factor & Setting & Orig. & Exp. & Fresh & Loss & Gain & Messages & Tokens \\
\midrule
Formation length & 0 episodes & 54.29 & 54.27 & 53.39 & 0.02 & 0.88 & $-$1.95 & $-$0.03 \\
 & 1 episode & 62.00 & 59.39 & 57.54 & 2.61 & 1.85 & 11.19 & 19.28 \\
 & 5 episodes (default) & 70.30 & 57.82 & 60.29 & 12.49 & $-$2.48 & 62.75 & 84.43 \\
 & 10 episodes & 74.67 & 64.26 & 61.72 & 10.41 & 2.54 & 50.83 & 73.43 \\
\midrule
Model family & Gemini $\rightarrow$ Gemini & 74.25 & 65.06 & 66.61 & 9.19 & $-$1.55 & 40.84 & 60.70 \\
 & DeepSeek $\rightarrow$ DeepSeek & 68.60 & 57.73 & 59.04 & 10.87 & $-$1.31 & 44.11 & 61.40 \\
 & GPT $\rightarrow$ Gemini & 70.30 & 63.75 & 61.90 & 6.56 & 1.85 & 44.34 & 71.72 \\
 & DeepSeek $\rightarrow$ Gemini & 68.60 & 60.00 & 63.08 & 8.60 & $-$3.08 & 40.81 & 65.21 \\
\midrule
Temperature & $T = 0.0$ & 69.86 & 61.86 & 59.90 & 8.00 & 1.96 & 37.43 & 56.40 \\
 & $T = 1.0$ & 68.25 & 60.26 & 56.23 & 7.99 & 4.04 & 45.28 & 67.78 \\
\midrule
Planning team & 2 members & 71.51 & 61.70 & 58.80 & 9.81 & 2.90 & 25.62 & 43.79 \\
 & 3 members & 72.97 & 66.62 & 62.37 & 6.34 & 4.25 & 13.51 & 30.59 \\
 & 5 members & 71.92 & 68.51 & 65.98 & 3.41 & 2.53 & 2.34 & 17.95 \\
\bottomrule
\end{tabular*}

\end{table*}

\paragraph{Agents and evaluation} The default configuration uses GPT-5.6 Luna at temperature $T = 0.7$, five formation episodes and a 1,000-token notebook budget. The model-family comparison also uses Gemini 3.8 Flash and DeepSeek V4 Flash, and a newcomer that runs model $Y$ carries the notebook it wrote in a donor team of model $Y$. In Collab-Overcooked and CREW-Wildfire each branch is evaluated on five episodes per repetition, about 4,000 evaluation episodes in all. In MedQA each branch receives 1,000 test-question evaluations per repetition, drawn from the 1,273-question test split.

\paragraph{Study design} Two kinds of study answer different questions, and unless stated otherwise both use the default configuration. A breadth screen checks the direction of both contrasts beyond one task family. It runs the three branches in all five settings, ten repetitions each, and in Collab-Overcooked replaces the initiator in five repetitions and the responder in the other five. Depth studies on Collab-Overcooked, whose two seats differ in structure (Figure~\ref{fig:swap}, bottom left), estimate the contrasts with intervals and locate their source. The seat study replays both seats from every formation snapshot, with ten receiver--donor pairs on each of L1 to L3 and a replication with twenty further pairs on L3. The content study reuses the seat study's L3 snapshots. The configuration study varies formation length, model family, temperature and team organization on L3 and replaces each seat in five of ten repetitions, so its gains average the two seats and we read it mainly through the loss. Each repetition runs the whole protocol, from formation to evaluation, and is the experimental unit: contrasts are computed within a repetition and averaged over repetitions. Intervals are 95\% Student-$t$ intervals of the mean, and $k$/$n$ counts the $k$ of $n$ repetitions that share the sign of the mean. Communication premiums are reported as point values. Scoring rules, prompts, notebook compositions and repetition-level summaries are in the supplementary material.

The results answer four questions in turn: whether the usual comparison misses a cost (Section~\ref{sec:screen}), whether that cost comes from the agents or from the history a team builds (Section~\ref{sec:formation}), where in the history it lies and for which seat (Sections~\ref{sec:seat} and~\ref{sec:content}), and whether the way a team is organized can reduce it (Section~\ref{sec:planning}).

\subsection{The usual comparison approves a costly replacement}\label{sec:screen}
Table~\ref{tab:breadth} gives the screen's verdict under each baseline. Against a fresh newcomer, imported experience helps in all five settings, by 1.96 to 6.16 points. Against the original team, the same replacement falls short in all five, by 1.25 to 7.05 points, and the team needs 5.31\% to 34.33\% more messages and 4.16\% to 33.17\% more tokens per unit of progress. Judged by the first comparison alone, the transfer would pass in every setting. Outside cooking, the loss is 2.63 points in CREW-Wildfire, with a 9.08\% message premium, and 2.29 points in MedQA, where the team discusses instead of acting and the premiums are smallest, 5.31\% for messages and 4.16\% for tokens. Within Collab-Overcooked both costs grow with the level, the loss from 1.25 points on L1 to 7.05 on L3 and the message premium from 15.45\% to 34.33\%. On L3 the pooled gain of $+6.16$ combines two seats whose gains have opposite signs, $-2.70$ \CI{$-8.32$}{2.92} when the initiator is replaced (4/5 negative) and $+15.02$ \CI{8.61}{21.43} when the responder is (5/5). Section~\ref{sec:seat} measures each seat on every snapshot.

\subsection{Formation creates the loss, and changing the agents leaves it}\label{sec:formation}
\begin{figure*}
\centering
\includegraphics[width=0.933\textwidth,trim={0 3pt 0 2pt},clip]{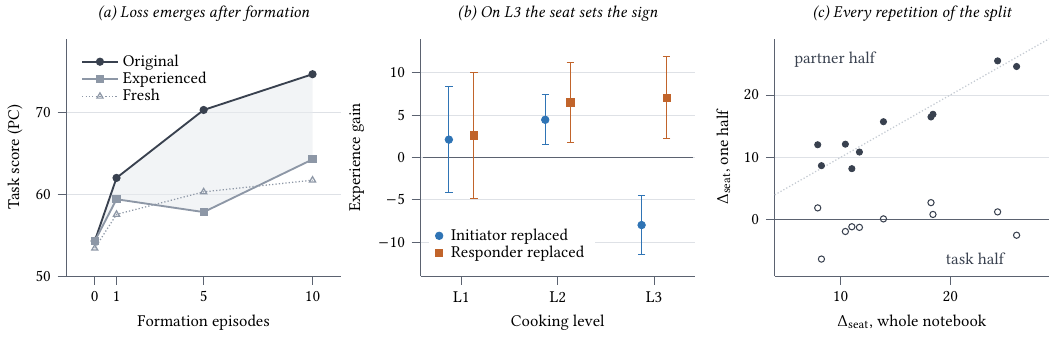}
\caption{The argument in three steps. (a) Task score against formation length on L3, pooled over the replaced seats, with the loss shaded. (b) Gain E $-$ F by level and replaced seat in the default configuration, with 95\% intervals over ten repetitions. (c) Seat asymmetry per repetition: the partner half (filled) tracks the whole notebook and the task half (open) does not. The dotted line marks equality.}
\Description{Three panels. (a) Line plot of task score against formation episodes (0, 1, 5, 10) on L3 for the Original, Experienced and Fresh branches, with the continuity loss shaded. (b) Experience gain with 95\% intervals for L1, L2 and L3 when the initiator or the responder is replaced. (c) Scatter plot of the per-repetition seat asymmetry of the whole notebook against that of the partner half and of the task half.}
\label{fig:argument}
\end{figure*}

\paragraph{Formation length} We vary the number of formation episodes on L3 (0, 1, 5 or 10) and hold everything else fixed (Table~\ref{tab:config}, Figure~\ref{fig:argument}a). With no formation every notebook is empty, so Experienced and Fresh are the same condition and differ from Original only in the notice. Original and Experienced then score 54.29 and 54.27, a loss of 0.02 points \CI{$-2.57$}{2.61}. Before anything is written down, neither the swap nor the notice produces a detectable loss. The loss appears once the team has worked together: 2.61 points after one formation episode (8/10), 12.49 after five (8/10) and 10.41 after ten (10/10). It is positive in all twelve rows of Table~\ref{tab:config} that include formation. Figure~\ref{fig:argument}a pools the two seats, as the configuration study does, and at five episodes its Experienced curve dips below Fresh. Section~\ref{sec:seat} shows that the two seats' gains have opposite signs there.

\paragraph{What formation builds} Ten formation episodes lift the L3 team from 53.39, the unformed team with a fresh newcomer, to 74.67. When one member is replaced, the retained teammate's notebook keeps 8.33 of these 21.28 points, the imported notebook recovers 2.54, and 10.41 leave with the replaced member. Even with an experienced replacement, the team loses about half of what formation built. The imported notebook brings back about an eighth, and that eighth is the whole of what the usual baseline measures.

\paragraph{Changing the agents} If the loss were a quirk of one model or one temperature, changing them should remove it. We change which model family occupies the seats, writing $X \rightarrow Y$ for receiving model $X$ and incoming model $Y$, and the temperature at which the whole team decodes. Every changed model pairing and temperature leaves a loss whose interval excludes zero (Table~\ref{tab:config}, with intervals in the supplementary material): from 6.56 \CI{2.85}{10.27} for GPT $\rightarrow$ Gemini to 10.87 \CI{9.10}{12.65} for DeepSeek $\rightarrow$ DeepSeek across model families, and 8.00 \CI{4.20}{11.80} and 7.99 \CI{6.32}{9.66} at $T = 0.0$ and $T = 1.0$. The cross-model rows also show that a notebook has no fixed value. The same Gemini newcomers, carrying the same donor notebooks, have a pooled gain of $+1.85$ \CI{$-1.08$}{4.78} in GPT teams and $-3.08$ \CI{$-4.46$}{$-1.70$} in DeepSeek teams, where the gain is negative in all ten repetitions, five per seat. The two intervals do not overlap. What a notebook is worth depends on the team that receives it.

\subsection{The seat can reverse the sign of imported experience}\label{sec:seat}
\paragraph{Design} To measure each seat on its own, the seat study branches every formed team twice from one snapshot, once replacing the initiator and once the responder, and compares both replacements with the same Original evaluation (Table~\ref{tab:seat}).

\paragraph{Results} On L1 and L2 we detect no difference between the seats. The mean gain is positive in both seats, and the seat asymmetry is 0.50 \CI{$-3.38$}{4.39} on L1 and 2.04 \CI{$-2.82$}{6.90} on L2. On these levels the initiator's reference sequence is only slightly longer than the responder's, a point that Section~\ref{sec:discussion} returns to. L3 separates the seats (Figure~\ref{fig:argument}b). Replacing the initiator, the harder seat in this benchmark by its authors' account \cite{sun2025collabovercooked}, costs 16.26 points against 6.25 for the responder, and the imported notebook changes sign. Relative to a fresh newcomer, the experienced initiator scores 7.98 points lower and the experienced responder 7.05 points higher, so that $\Dseat = 15.03$ \CI{10.41}{19.65} (10/10). A replication with twenty further receiver--donor pairs gives the same pattern (Table~\ref{tab:contrasts} and supplementary material): losses of 15.60 and 6.90 points, gains of $-7.10$ and $+6.40$, and $\Dseat = 13.50$ \CI{10.05}{16.95}, positive in 19 of the 20 pairs.

The team's own history helps in either seat, while a donor's history hurts in one of them. Against a fresh newcomer, the incumbent initiator is worth 8.28 points and the incumbent responder 13.30. Two thirds of $\Dseat$, 10.02 points \CI{6.06}{13.98} (9/10), come from the Experienced branches and so equal the difference between the two seats' losses. The remaining third comes from the Fresh branches, 62.02 against 57.00, and is the amount by which the incumbent responder's value exceeds the incumbent initiator's. The same donor teams thus supply an asset in one seat and a liability in the other, and a claim that experience transfers has to name the seat. The receiving team matters as well: in DeepSeek teams, the Gemini newcomers' notebooks fell short of an empty one in both seats (Section~\ref{sec:formation}).

Communication points the same way. Replacing the L3 initiator costs 70.45\% more messages and 96.08\% more tokens per unit of progress, against 28.19\% and 45.89\% for the responder. Per episode, the team sends 25.3\% more messages when the initiator is replaced and 16.9\% more when the responder is, so each premium combines more talk with less progress.

\begin{table*}
\caption{Seat study in the default configuration. On L3 the gain changes sign with the replaced seat. Each row is a mean over ten repetitions with the named seat replaced, and both seats share one Original.}
\label{tab:seat}
\begin{tabular*}{\textwidth}{@{\extracolsep{\fill}}llrrrrrrr@{}}
\toprule
& & \multicolumn{5}{c}{Task score (PC)} & \multicolumn{2}{c}{Premium (\%)} \\
\cmidrule(lr){3-7}\cmidrule(l){8-9}
Level & Replaced seat & Orig. & Exp. & Fresh & Loss & Gain & Messages & Tokens \\
\midrule
L1 & Initiator & 91.35 & 88.60 & 86.49 & 2.75 & 2.10 & 23.25 & 28.85 \\
 & Responder & 91.35 & 89.24 & 86.64 & 2.11 & 2.61 & 24.34 & 29.32 \\
\midrule
L2 & Initiator & 85.08 & 81.32 & 76.88 & 3.76 & 4.44 & 19.40 & 31.77 \\
 & Responder & 85.08 & 80.57 & 74.09 & 4.51 & 6.48 & 22.71 & 34.96 \\
\midrule
L3 & Initiator & 70.30 & 54.04 & 62.02 & 16.26 & $-$7.98 & 70.45 & 96.08 \\
 & Responder & 70.30 & 64.06 & 57.00 & 6.25 & 7.05 & 28.19 & 45.89 \\
\bottomrule
\end{tabular*}

\vspace{10pt}
\caption{Notebook content on L3. Rewording costs almost nothing, and the partner half acts like the whole notebook. Change is measured against the first row of each block and the token premium against the Original.}
\label{tab:content}
\begin{tabular*}{\textwidth}{@{\extracolsep{\fill}}llrrrrrr@{}}
\toprule
& & \multicolumn{2}{c}{Task score (PC)} & \multicolumn{2}{c}{Change} & \multicolumn{2}{c}{Token premium (\%)} \\
\cmidrule(lr){3-4}\cmidrule(lr){5-6}\cmidrule(l){7-8}
History in the seat & Notebook the seat carries & Init. & Resp. & Init. & Resp. & Init. & Resp. \\
\midrule
Its own & Kept as written & 70.30 & 70.30 &  &  & 0.00 & 0.00 \\
 & Reworded, same content & 69.78 & 70.35 & $-$0.52 & 0.05 & 4.04 & $-$0.40 \\
\midrule
A newcomer's & None & 62.02 & 57.00 &  &  & 70.26 & 96.68 \\
 & Task-procedural half & 63.49 & 57.88 & 1.47 & 0.88 & 50.33 & 79.84 \\
 & Partner-referring half & 54.11 & 64.20 & $-$7.91 & 7.19 & 79.78 & 42.11 \\
 & Whole notebook & 54.04 & 64.06 & $-$7.98 & 7.05 & 96.08 & 45.89 \\
\bottomrule
\end{tabular*}

\end{table*}

\subsection{Lines about the partner carry the seat effect}\label{sec:content}
\paragraph{Design} Section~\ref{sec:seat} leaves open which part of the donor notebook carries the asymmetry, and it does not exclude a simpler explanation: any notebook that its reader did not write may act as an unfamiliar prompt. Two further interventions on the same L3 snapshots address both points. The first, a paraphrase control, keeps the incumbent in its seat with its own notebook, rewritten by GPT-5.6 Terra with the instruction to keep every fact, instruction and reference, to change only wording and, within a section, sentence order, and to stay within the same length budget. Everything else is as in the Original branch, with no notice, so only the surface form of one notebook differs. The second intervention splits each donor notebook into content about the task and content about the partner. A line is \emph{partner-referring} if its main assertion concerns the other agent: what it will do, needs to be told or expects in return, including agreed handoffs, whether or not this is specific to one partner. A line stating task facts or procedures, such as recipes, timings and action orders, is \emph{task-procedural}, and a line touching both takes the label of its main assertion. The staging note quoted in Section~\ref{sec:intro}, from a donor initiator's notebook, is partner-referring. Two language models other than the one that wrote the notebooks, Gemini 3.5 Flash-Lite and DeepSeek V4 Flash, labelled each of the 141 lines of the twenty donor notebooks independently with these definitions. They agreed on 121 lines (85.8\%). GPT-5.6 Terra, given the same definitions, resolved the other 20, and three lines that fit neither class were dropped. The taxonomy was fixed before any half was transferred. In either role, task-procedural lines make up about 53\% of a notebook and partner-referring lines 44\% to 45\%, so the two halves are of similar length. Each half, the \emph{task half} and the \emph{partner half}, is then installed on its own in the replaced seat, with everything else, including the notice, as in the Experienced branch (Table~\ref{tab:content}).

\begin{table}[t]
\caption{Contrasts of Sections~\ref{sec:seat} and~\ref{sec:content} on L3, default configuration: mean, 95\% interval and $k$/$n$. The last row repeats $\Dseat$ in the replication with twenty further pairs.}
\label{tab:contrasts}
\begin{tabular*}{\columnwidth}{@{\extracolsep{\fill}}lrcr@{}}
\toprule
Contrast & Mean & 95\% interval & $k$/$n$ \\
\midrule
$\Dseat$, whole notebook & 15.03 & $[10.41, 19.65]$ & 10/10 \\
$\Dseat$, partner half & 15.10 & $[10.79, 19.42]$ & 10/10 \\
$\Dseat$, task half & $-0.59$ & $[-2.47, 1.28]$ & 5/10 \\
\midrule
Initiator gain, whole & $-7.98$ & $[-11.44, -4.51]$ & 9/10 \\
Initiator gain, partner half & $-7.91$ & $[-11.49, -4.33]$ & 10/10 \\
Initiator gain, task half & 1.47 & $[0.57, 2.37]$ & 8/10 \\
Responder gain, whole & 7.05 & $[2.20, 11.90]$ & 8/10 \\
Responder gain, partner half & 7.19 & $[2.44, 11.95]$ & 8/10 \\
Responder gain, task half & 0.88 & $[-0.30, 2.06]$ & 9/10 \\
\midrule
Rewording cost, initiator & 0.52 & $[-0.08, 1.13]$ & 8/10 \\
Rewording cost, responder & $-0.05$ & $[-0.67, 0.57]$ & 4/10 \\
\midrule
$\Dseat$, replication (20 pairs) & 13.50 & $[10.05, 16.95]$ & 19/20 \\
\bottomrule
\end{tabular*}
\end{table}

\paragraph{Results} Rewording costs almost nothing: 0.52 points in the initiator seat and $-0.05$ in the responder seat, against 16.26 and 6.25 for handing the seat to an experienced newcomer, with token premiums of 4.04\% and $-0.40$\% (Tables~\ref{tab:content} and~\ref{tab:contrasts}). No repetition loses more than 2.20 points, so unfamiliar wording accounts for almost none of the loss. The partner half reproduces the whole notebook's seat asymmetry (Figure~\ref{fig:argument}c): $\Dseat = 15.10$ against 15.03 (10/10). Per repetition the two values differ by 0.07 points on average \CI{$-1.45$}{1.60}, and they correlate at 0.94. The task half reproduces none of the asymmetry, $-0.59$, with five repetitions on each side of zero. Seat by seat, the partner half matches the whole notebook within 0.15 points, while the task half changes the score over a fresh newcomer by only 1.47 and 0.88 points. Installed alone, the task half beats the whole notebook by 9.45 points \CI{5.71}{13.19} in the initiator seat (10/10) and trails it by 6.17 \CI{1.66}{10.69} in the responder seat (8/10). Communication follows the same split. In the initiator seat the partner half raises the token premium more than the task half does, 79.78\% against 50.33\%, and in the responder seat less, 42.11\% against 79.84\%.

\paragraph{Interpretation} Because the donor team scored no higher during formation, its notebook could simply be worse than the receiver's. The halves show that it is not worse throughout. Both halves come from the same donor notebook, yet the task half changes the score little in either seat, and the partner half lifts a newcomer responder 7.19 points above an empty notebook. Only the partner half's gain changes sign with the seat. Task procedure transfers alike into either seat, while partner-referring content carries the whole asymmetry. The two kinds of content can be separated, so experience need not be discarded whole.

\subsection{Planning together narrows the loss}\label{sec:planning}
\paragraph{Design} Changing the agents left the loss in place (Section~\ref{sec:formation}), so we change how the team is organized. Members who only plan can help the team agree on who does what before the first handoff. The initiator and the responder remain the acting roles, and we add members who take part in planning only, sharing one interface and a team-wide budget of 128 planning tokens per round. The comparison changes three things together: the number of planners, each member's share of the budget and the number of retained notebooks. Planning exchanges count toward communication, and the two-member planning team, which uses the same interface, is the baseline.

\paragraph{Results} Growing the planning team from two members to three and then five raises Experienced PC from 61.70 to 66.62 and 68.51 and lowers the loss from 9.81 to 6.34 and 3.41 points (Table~\ref{tab:config}). The five-member loss is the lowest of any L3 configuration with five formation episodes. Paired within repetitions, the reduction is 6.40 points \CI{4.11}{8.68} (10/10). Original barely moves, from 71.51 to 71.92, so the narrowing comes from the Experienced branch. The communication premiums fall from 25.62\% to 2.34\% in messages and from 43.79\% to 17.95\% in tokens. A team in which more members plan with the newcomer before anyone acts loses less when one acting member is replaced, although a loss of 3.41 points remains.

\section{Discussion}\label{sec:discussion}
\paragraph{Who predicts whom} We read a partner-referring line as a prediction about a partner. Its value then depends on which seat has to make the prediction. On L3 the initiator holds the recipe and runs most of the plan, so it is the responder that predicts the initiator. A newcomer responder follows: its donor's record of how initiators act is a usable prior, and the retained initiator's actions correct it. A newcomer initiator leads: it acts on staging and handoff arrangements that the retained responder never agreed to, and the responder, which has not been told of any change, has no reason to question them. Hierarchical accounts of convention draw the same line between common ground with one partner and community-wide conventions, which serve as priors for new partners \cite{hawkins2023partners}. Because the recipe dictates how any initiator acts, a responder's record of initiators is close to a community-wide prior, whereas an initiator's staging and handoff arrangements are common ground with one partner. The staging note quoted in Section~\ref{sec:intro} illustrates the second case. Carried by a newcomer initiator, it would direct handoffs to counters [2,2] and [2,3], an arrangement that the new responder never made.

Four observations fit this account. The partner half carries the whole asymmetry, while the task half, which describes the world instead of a partner, transfers alike into either seat. With a fresh replacement, losing the responder, the member that predicts, costs the team more than losing the initiator, 13.30 against 8.28 points. The asymmetry is large on L3, where the initiator's reference sequence is 1.46 times the responder's, and undetectable on L1 and L2, where the ratio is 1.08 and 1.13. And the loss shrinks when more members plan with the newcomer before anyone acts. The account also predicts what further swaps should show: telling the retained teammate of the change, or letting it state the team's plan first, should reduce the loss under a newcomer initiator.

\paragraph{Implications for LLM teams} If the cost lies in arrangements that only the retained members share, a replacement should be followed by re-establishing those arrangements with the newcomer, and the planning study is consistent with this. Evaluation should take place in the receiving team, seat by seat, with communication reported beside the score, because every screened transfer passed when judged against a fresh newcomer. Task procedure can travel, as it did not hurt in either seat. Memory systems that pool or share experience \cite{chang2026memcollab,zhang2025gmemory}, and methods that filter memory before sharing it \cite{chang2026memcollab,tian2026masking}, would gain from marking what a memory says about partners, so that a replacement carries its task knowledge and brings partner-referring content only to seats where it helps. An agent could mark each line as it writes it, as our labelling did after the fact.

\paragraph{Scope and open questions} The swap measures a team at the handover, the point at which a replacement decision is made, on the tasks it formed on, with bounded written notebooks and a notice that only the newcomer receives. The depth studies use one benchmark, and the seat and content analyses one configuration. The same protocol can extend this scope directly. It can measure how fast the loss closes as the new team accumulates history, whether partner-referring content still misleads on held-out tasks, whether the seat asymmetry persists without the notice, and how the loss changes in teams with more acting members. Assigning receiver and donor at random would also show how much of the loss depends on which team scored higher during formation.

\section{Conclusion}
In the settings we studied, an experienced agent was not interchangeable with a member of an already formed team, even with the same role, the same model and as much formation behind it. Against an empty notebook, imported experience looked like a success in every screened setting. Compared with the team as it was, it fell short and cost more communication. On L3 the loss appeared only with formation and outlasted changes of model family and temperature. The value of a donor notebook depended on the seat and on the receiving team, and its lines about the partner carried the whole difference between seats. An agent should therefore be evaluated in its receiving team, by what the team achieves and the communication it spends, rather than certified as a drop-in replacement.

\clearpage

\clearpage
\makeatletter
\@ACM@balancefalse
\makeatother
\appendix
\setcounter{table}{0}
\setcounter{figure}{0}
\setcounter{equation}{0}
\renewcommand{\thetable}{S\arabic{table}}
\renewcommand{\theHtable}{appendix.\arabic{table}}
\renewcommand{\theHfigure}{appendix.\arabic{figure}}
\renewcommand{\theHequation}{appendix.\arabic{equation}}
\newcommand{\sO}{\mathrm{O}}
\newcommand{\sE}{\mathrm{E}}
\newcommand{\sF}{\mathrm{F}}
\newcommand{\sinit}{\mathrm{init}}
\newcommand{\sresp}{\mathrm{resp}}
\newcommand{\TES}{\mathrm{TES}}
\fvset{fontsize=\small,xleftmargin=0.8em}
\section{Experimental settings}\label{sec:s-settings}
\paragraph{Agents} Each agent is a fixed language model with a role instruction, an action interface and a private notebook, and no model weights are updated. In the embodied settings an agent returns, at each decision, one legal action, a public message that may be empty, and a private rationale. Only the message is shared. All agent prompts are in Appendix~\ref{sec:s-prompts}.

\paragraph{Models} The default model is GPT-5.6 Luna at temperature $T = 0.7$, with reasoning effort set to none so that the temperature applies. The model-family comparison uses Gemini 3.8 Flash and DeepSeek V4 Flash. A newcomer that runs model $Y$ carries the notebook it wrote in a donor team of model $Y$, and the GPT $\rightarrow$ Gemini and DeepSeek $\rightarrow$ Gemini rows use the same Gemini newcomers with the same donor notebooks. The temperature comparison sets $T = 0.0$ or $T = 1.0$ for the whole team.

\paragraph{Notebook} Each agent rewrites its notebook between formation episodes from its experience and the outcome feedback, as a cumulative notebook of concise numbered notes, one main point per note, that keeps earlier lessons unless new evidence contradicts them. No number of lines and no mix of content is prescribed (prompt in Appendix~\ref{sec:s-notebook-prompt}). The budget is 1,000 tokens in all experiments.

\paragraph{Formation} Two teams start from empty notebooks and play the same formation sequence: five episodes by default, and 0, 1, 5 or 10 when formation length varies. The team with the higher mean formation score becomes the receiver and the other the donor, with ties broken by a fixed index.

\paragraph{Evaluation} Notebooks are frozen and reloaded at every evaluation episode, and all branches of a comparison share one task list. In the embodied settings each branch is evaluated on five episodes per repetition. In MedQA each branch receives 1,000 question evaluations per repetition, 10,000 in all, drawn from the 1,273 questions of the test split, so questions recur across repetitions.

\paragraph{Newcomer notice} The newcomer, experienced or fresh, receives in every evaluation episode the line \emph{``This is your first episode with this team in the $\langle$role$\rangle$ seat.''} The incumbent and the retained teammates receive no such line.

\paragraph{Collab-Overcooked} Levels L1 to L3 \cite{sun2025collabovercooked}. The initiator alone holds the recipe and operates the pot, oven and serving, and the responder prepares ingredients and dishes and hands them over. Each level has five recipes, taken in a fixed order with a starting point that rotates across repetitions, and evaluation covers the same five recipes. The initiator's reference action sequence is on average 1.08, 1.13 and 1.46 times as long as the responder's on L1, L2 and L3 (34.8 against 23.8 actions on L3). Episodes are capped at 1.5$\times$ the reference completion time (scoring in Appendix~\ref{sec:s-overcooked}).

\paragraph{CREW-Wildfire} Three acting agents with identical capabilities, one of whom is replaced at random, in a sparse tree-cutting task with six target cells and twelve joint decision rounds \cite{hyun2025crewwildfire} (scoring in Appendix~\ref{sec:s-wildfire}).

\paragraph{MedQA} Four-option USMLE questions from MedQA \cite{jin2021medqa}, answered by the four-agent ColaCare topology \cite{wang2025colacare}, one of the multiagent frameworks of MedAgentBoard \cite{zhu2025medagentboard}: three specialists and a coordinator, stopping at agreement or after three rounds. Only specialist seats are replaced. Each formation episode is one training question, and evaluation uses test questions (scoring in Appendix~\ref{sec:s-medqa}, prompts in Appendix~\ref{sec:s-medqa-prompts}).

\paragraph{Planning variants} The initiator and the responder remain the acting roles. Added members take part in planning only, through one shared interface with a team-wide budget of 128 planning tokens per round, shared among the members (Appendix~\ref{sec:s-comm}, prompt in Appendix~\ref{sec:s-planning-prompts}). Teams have 2, 3 or 5 members.

\paragraph{Breadth screen} The screen runs all three branches in each of the five settings in the default configuration, with ten independent repetitions per setting, separate from those of the depth studies. In Collab-Overcooked the initiator is replaced in five repetitions and the responder in the other five. Branch means are reported, and on L3 also the split by seat (Table~\ref{tab:s-screen}).

\paragraph{Seat study and replication} The seat study and its replication use the default configuration and replay both seats from each formation snapshot, sharing one Original evaluation. The seat study runs ten receiver--donor pairs on each of L1 to L3, and the replication twenty further pairs on L3 (Table~\ref{tab:s-second}).

\section{Protocol and contrasts}\label{sec:s-protocol}
\subsection{Branches}
Two teams form independently in the same task setting and are frozen. The team with the higher mean score over its own formation episodes is labelled the receiver $R$ and the other the donor $D$. Ties, including the zero-formation case, are broken by a fixed index. The rule uses formation records only, and the donor's notebooks are transferred without selection or editing. With $m^R_j$ and $m^D_j$ the notebooks at role $j$ and $m^R_{-j}$ the notebooks of the receiver's other members, the three branches are
\begin{align*}
\mathcal{M}_{\mathrm{O}} &= (m^R_{-j}, m^R_j),\\
\mathcal{M}_{\mathrm{E}} &= (m^R_{-j}, m^D_j),\\
\mathcal{M}_{\mathrm{F}} &= (m^R_{-j}, \varnothing).
\end{align*}
Outside the model-family comparison the newcomer runs the incumbent's model. The newcomer, experienced or fresh, receives the notice, and the incumbent in the Original branch does not.

\subsection{Replacement contrasts and the seat asymmetry}\label{sec:s-contrasts}
Contrasts are computed within each repetition $b = 1, \ldots, B$, with $B = 10$, or $B = 20$ in the replication. Let $S^s_{bc}$ be the score of branch $c \in \{\sO,\sE,\sF\}$ in repetition $b$ when seat $s \in \{\sinit,\sresp\}$ is replaced, the mean over that repetition's evaluation episodes (Appendix~\ref{sec:s-units}). The seat study and its replication evaluate both replacements from each snapshot and share one Original evaluation, $S^{\sinit}_{b\sO} = S^{\sresp}_{b\sO} = S_{b\sO}$. The continuity loss and the experience gain are the signed differences
\begin{equation*}
D^s_b = S_{b\sO} - S^s_{b\sE}, \qquad G^s_b = S^s_{b\sE} - S^s_{b\sF},
\end{equation*}
which the paper writes as $D_S$ and $G_S$, with $S_c$ for the score of one snapshot. The seat asymmetry is
\begin{align*}
\Delta_{\mathrm{seat},b} &= G^{\sresp}_b - G^{\sinit}_b\\
&= \underbrace{\bigl(S^{\sresp}_{b\sE} - S^{\sinit}_{b\sE}\bigr)}_{\text{Experienced term}}
+ \underbrace{\bigl(S^{\sinit}_{b\sF} - S^{\sresp}_{b\sF}\bigr)}_{\text{Fresh term}} .
\end{align*}
Because the Original branch is shared, the Experienced term also equals $D^{\sinit}_b - D^{\sresp}_b$, the difference between the two seats' losses. These contrasts are linear in the scores, so their means over the repetitions equal the same contrasts of the branch means $\bar S^s_c = B^{-1} \sum_b S^s_{bc}$, which the tables of the paper report.

\paragraph{What the two terms compare} Replacing the initiator runs the pair $(m^D_{\sinit}, m^R_{\sresp})$ and gives the notice to the initiator. Replacing the responder runs $(m^R_{\sinit}, m^D_{\sresp})$ and gives the notice to the responder. The Experienced term therefore compares two teams that differ in which seat holds the newcomer with its notice and in which seat holds a notebook of the team that scored higher during formation. The Fresh term compares two teams that differ in whose own history is missing, and it involves no donor notebook. Neither term involves the Original evaluation. The loss $D^s_b$ compares the incumbent with this particular replacement, so it includes any difference in notebook quality between the team that scored higher during formation and the team that did not.

\paragraph{Configuration and cross-model rows} Each row of the configuration study (Table~2 of the paper) replaces the initiator in five repetitions and the responder in the other five, and a repetition keeps its seat for both newcomer branches and every comparison, so a row's gain averages two seats. The five-episode row reuses the seat study's snapshots with one seat per repetition (Appendix~\ref{sec:s-res-formation}). In the cross-model rows the Experienced and Fresh newcomers run the same incoming model, and the GPT $\rightarrow$ Gemini and DeepSeek $\rightarrow$ Gemini rows reuse the same Gemini donor notebooks. GPT $\rightarrow$ Gemini shares its receiving teams and Original evaluation with the five-episode row, and DeepSeek $\rightarrow$ Gemini shares them with DeepSeek $\rightarrow$ DeepSeek. Scoring and aggregation are the same across the model, temperature and formation-length variants.

\subsection{Notebook and planning controls}\label{sec:s-controls}
\paragraph{Notebook halves} For each half $a \in \{\mathrm{task},\mathrm{partner}\}$ of the donor notebook (Appendix~\ref{sec:s-taxonomy}), let $S^s_{b,a}$ be the score when that half is installed alone in seat $s$. The half's gain, its difference from the whole notebook and its seat asymmetry are
\begin{gather*}
G^s_{b,a} = S^s_{b,a} - S^s_{b\sF}, \qquad H^s_{b,a} = S^s_{b,a} - S^s_{b\sE},\\
\Delta_{\mathrm{seat},b,a} = G^{\sresp}_{b,a} - G^{\sinit}_{b,a}.
\end{gather*}
A half keeps its original lines in their original order, with the dropped lines omitted. It shares the Fresh baseline of its seat and the notice, model and budgets of the Experienced branch, and no length correction is applied to scores. The partner half is compared with the whole notebook through the paired difference $\Delta_{\mathrm{seat},b,\mathrm{partner}} - \Delta_{\mathrm{seat},b}$ and through the Pearson correlation of $x_b = \Delta_{\mathrm{seat},b}$ and $y_b = \Delta_{\mathrm{seat},b,\mathrm{partner}}$ over the ten repetitions, with means $\bar x$ and $\bar y$,
\begin{equation*}
r = \frac{\sum_b (x_b - \bar x)(y_b - \bar y)}{\sqrt{\sum_b (x_b - \bar x)^2 \, \sum_b (y_b - \bar y)^2}}.
\end{equation*}

\paragraph{Paraphrase} The incumbent keeps its seat with its own reworded notebook and receives no notice (Appendix~\ref{sec:s-paraphrase}). With $S^s_{b,\mathrm{para}}$ the score of this branch, the cost of rewording is
\begin{equation*}
W^s_b = S_{b\sO} - S^s_{b,\mathrm{para}},
\end{equation*}
and the change in the own-notebook block of Table~4 of the paper is $-W^s_b$. Its token premium is that of Appendix~\ref{sec:s-comm}, with the paraphrase branch in place of Experienced.

\paragraph{Planning width} Teams of two, three and five members keep the initiator and the responder as the acting roles and plan through the shared interface. The two-member planning team is a baseline separate from the native two-agent setting, and each width $w$ has its own branch scores, costs and premiums. With $C^{q,(w)}_{b\sE}$ the Experienced cost of measure $q$ (Appendix~\ref{sec:s-comm}) and $D^{(w)}_b$ the loss at width $w$, both for the seat replaced in repetition $b$, the paired change in Experienced cost from two members to five, in percent, and the paired reduction in loss, in points, are
\begin{equation*}
\eta^q_b = 100\left(\frac{C^{q,(5)}_{b\sE}}{C^{q,(2)}_{b\sE}} - 1\right), \qquad z_b = D^{(2)}_b - D^{(5)}_b.
\end{equation*}
Table~\ref{tab:s-summary} summarizes both over the ten repetitions.

\section{Scoring and aggregation}\label{sec:s-scoring}
This appendix specifies the task scores, progress measures, communication costs and statistical summaries, including the conventions we adopt where the protocol leaves a choice open.

\subsection{Evaluation units}\label{sec:s-units}
Let $b = 1, \ldots, B$ index independent formation repetitions ($B = 10$, or 20 in the replication), $c \in \{\sO,\sE,\sF\}$ the Original, Experienced and Fresh branches, and $e$ an evaluation episode. Each embodied branch has five evaluation episodes per repetition. A MedQA episode is one test question, and each MedQA branch has 1,000 evaluations per repetition. Evaluation counts are equal across repetitions. All branches of a comparison use the same task inputs, and each episode reloads its branch's frozen notebooks. With episode score $S_{bce}$ and $N_b$ evaluated episodes, the repetition score and the branch mean are
\begin{equation*}
S_{bc} = \frac{1}{N_b} \sum_e S_{bce}, \qquad \bar S_c = \frac{1}{B} \sum_b S_{bc}.
\end{equation*}
Scores lie on a 0--100 scale and are compared within a task setting, and progress keeps the task's native units.

\subsection{Collab-Overcooked}\label{sec:s-overcooked}
\paragraph{Executed trace} For acting role $i \in \{\sinit,\sresp\}$, let $h_i$ be its executed interaction trace, of length $n_i = |h_i|$. The trace lists the accepted environment interactions in execution order, with the benchmark's canonical action names and arguments. Movement, waiting, messages and unexecuted actions are excluded, while accepted interactions that are irrelevant to the recipe remain. A processing-start event is recorded when its interaction is accepted, and the final delivery is included.

\paragraph{Reference matching} Let $g_{ir}$ be candidate reference trajectory $r$ for role $i$, of length $\ell_{ir}$. Its matched-prefix length is
\begin{equation*}
d_{ir} = \max\bigl\{ d : (g_{ir,1}, \ldots, g_{ir,d}) \text{ is a subsequence of } h_i \bigr\},
\end{equation*}
where a subsequence keeps the order of $h_i$ but need not be contiguous. Matching thus starts at the first reference action and cannot skip a reference step, while extra executed actions may occur between matched steps.

\paragraph{Task score} The benchmark's Trajectory Efficiency Score \cite{sun2025collabovercooked} is
\begin{equation*}
\TES_{ir} = \frac{(1 + \beta^2)\, d_{ir}}{\ell_{ir} + \beta^2 n_i}, \qquad \beta = 0.95.
\end{equation*}
For each role the scorer selects the reference $r^*_i$ with the largest TES, breaking score ties by the larger matched-prefix length and then by a fixed reference order. We write $\TES_i = \TES_{ir^*_i}$ and $d_i = d_{ir^*_i}$. Progress Completeness (PC) on the 0--100 scale is
\begin{equation*}
S_{bce} = 50\,\bigl(\TES_{\sinit} + \TES_{\sresp}\bigr),
\end{equation*}
so the two acting roles have equal weight. Planning-only members do not enter this average.

\paragraph{Progress and termination} With the same selected references,
\begin{equation*}
P_{bce} = d_{\sinit} + d_{\sresp}.
\end{equation*}
Progress counts matched interactions without normalization. A redundant interaction that leaves the selected prefix unchanged does not increase progress, although its contribution to the trace length can reduce TES. An episode ends at task completion or at 1.5 times the reference completion time $\tau_{\mathrm{ref}}$, implemented as the discrete cap $\lfloor 1.5\, \tau_{\mathrm{ref}} \rfloor$ in environment time units. Scores are computed from the complete trace at termination. An incomplete episode keeps its achieved score and progress, and an empty trace has zero TES and zero progress.

\subsection{CREW-Wildfire}\label{sec:s-wildfire}
\paragraph{Native score} The sparse tree-cutting task \cite{hyun2025crewwildfire} has three acting agents and six target cells. With $n_j(t)$ the number of trees remaining in target cell $j$, the native raw score is
\begin{equation*}
V_t = \sum_{j=1}^{6} \bigl(3 - n_j(t)\bigr),
\end{equation*}
which reaches its maximum of 18 when every target cell is cleared, $n_j(t) = 0$. A target cell can start with one, two or three trees, so the initial score $V_0$ includes credit for trees absent at initialization.

\paragraph{Progress and score} Progress excludes this initial credit,
\begin{equation*}
P_{bce} = V_{\mathrm{end}} - V_0,
\end{equation*}
where $V_{\mathrm{end}}$ is the score at the terminal state, so $P_{bce}$ is the number of target trees cut during the episode. The task score normalizes completion by the work remaining at initialization,
\begin{equation*}
S_{bce} = 100\, \frac{V_{\mathrm{end}} - V_0}{18 - V_0},
\end{equation*}
where $18 - V_0 = \sum_j n_j(0)$ is the initial number of target trees. Each episode is normalized before scores are averaged, and comparison branches share the same initial target state and hence the same denominator. Removing trees outside the target cells does not increase progress.

\paragraph{Decision rounds} Evaluation is capped at twelve joint decision rounds. Our adapter treats a round as the execution phase of the three selected high-level actions: it advances the underlying transitions that they require, and an agent whose action has completed stays idle until the next round. Native termination and execution failures are handled by the executor, and the terminal state is sampled after the final execution phase. Underlying transitions create no additional decision rounds or messages.

\subsection{MedQA}\label{sec:s-medqa}
\paragraph{Team answer} Three specialists and a coordinator discuss each question, and only specialist seats are replaced. The controller follows the ColaCare sequence \cite{wang2025colacare}: each round has three specialist analyses, one coordinator synthesis, three specialist reviews and one coordinator decision (prompts in Appendix~\ref{sec:s-medqa-prompts}). It stops when all three reviews agree with the synthesis, or after three rounds, and takes the coordinator's decision from the terminating round.

\paragraph{Accuracy} Let $a_{bce}$ be the final option label and $a^*_{bce}$ the reference label. After deterministic option-label parsing,
\begin{equation*}
I_{bce} = \mathbf{1}[a_{bce} = a^*_{bce}], \qquad S_{bce} = 100\, I_{bce}, \qquad P_{bce} = I_{bce}.
\end{equation*}
The parser trims surrounding whitespace and normalizes label case against the supplied option set, and a missing, invalid or ambiguous label is incorrect. Agreement does not contribute to accuracy, and reference answers are available only to the scorer. The repetition score is therefore the percentage of questions answered correctly. Formation uses training questions and evaluation uses test questions.

\subsection{Shared communication}\label{sec:s-comm}
\paragraph{Messages} Each nonempty contribution actually published to the team counts once, and a broadcast counts once regardless of its number of recipients. Reading a contribution again in a later prompt does not create a new message, while sending it again does. Totals include unsuccessful episodes. Private reasoning, notebooks, notebook updates, environment observations and fixed prompt instructions are excluded. In the MedQA interface, analyses, syntheses, reviews and decisions are published to the shared discussion, so a complete round contributes eight messages. The count applies to published contributions, not to model calls or controller logs.

\paragraph{Tokens} Token counts do not use the models' own tokenizers. Every published contribution, whichever model produced it, is tokenized separately with the same tokenizer, the o200k\_base encoding of tiktoken (\url{https://github.com/openai/tiktoken}), and its count is added to the episode total. Embodied and planning messages use their published text. MedQA analyses, syntheses and decisions use the answer label followed by the public explanation, separated by a newline. Reviews use the Boolean agreement value, the answer label and the public reason, separated by newlines. JSON keys, fixed routing metadata and repeated prompt wrappers are excluded.

\paragraph{Planning budget} The shared planning interface has a team-wide budget of 128 tokens per round, that is, per planning phase (Appendix~\ref{sec:s-planning-prompts}). The budget is divided equally, with integer remainders assigned in a fixed member order: 64/64, 43/43/42 and 26/26/26/25/25 for two, three and five members. Only published tokens count toward communication. Planning exchanges and acting-role messages both enter the totals, and unused budget does not.

\paragraph{Cost per unit of progress} Let $M_{bce}$ and $K_{bce}$ be the episode's shared-message and shared-token totals. Costs are pooled within a repetition before division,
\begin{equation*}
C^{\mathrm{msg}}_{bc} = \frac{\sum_e M_{bce}}{\sum_e P_{bce}}, \qquad C^{\mathrm{tok}}_{bc} = \frac{\sum_e K_{bce}}{\sum_e P_{bce}},
\end{equation*}
and the repetition-level ratios are averaged with equal weight into branch-level costs,
\begin{equation*}
C^q_c = \frac{1}{B} \sum_b C^q_{bc}, \qquad q \in \{\mathrm{msg},\mathrm{tok}\}.
\end{equation*}
For either measure, the communication premium, in percent, is
\begin{equation*}
R^q_C = 100\left(\frac{C^q_{\sE}}{C^q_{\sO}} - 1\right),
\end{equation*}
computed from the unrounded branch-level costs. The paper writes $C_c$ and $R_C$ without $q$.

\paragraph{Zero progress} An episode with zero progress still contributes its messages and tokens to the repetition totals. A repetition with zero total progress would have an undefined cost. It would be kept, and its branch cost and premium reported as unavailable.

\subsection{Statistical summaries}\label{sec:s-stats}
The repetition is the experimental unit, and no repetition is dropped from any summary other than the robustness check in Appendix~\ref{sec:s-res-formation}. For a repetition-level contrast $x_b$ over $B$ repetitions, the mean, the sample standard deviation and the 95\% Student-$t$ interval are
\begin{gather*}
\bar x = \frac{1}{B} \sum_b x_b, \qquad s_x = \sqrt{\frac{\sum_b (x_b - \bar x)^2}{B - 1}},\\
\bar x \pm t_{0.975,\,B-1}\, \frac{s_x}{\sqrt{B}},
\end{gather*}
with $t_{0.975,\,9} = 2.262157$ for ten repetitions, $t_{0.975,\,19} = 2.093024$ for the twenty pairs of the replication, $t_{0.975,\,4} = 2.776445$ for the five repetitions per seat in the split of the breadth screen and $t_{0.975,\,6} = 2.446912$ for the seven repetitions of the robustness check.
The number of repetitions that share the sign of the mean,
\begin{equation*}
k = \sum_b \mathbf{1}[x_b\, \bar x > 0],
\end{equation*}
is reported as $k/B$. A zero contrast does not count toward $k$. Communication premiums are ratios of branch-level costs and are reported as point values, and the breadth screen reports branch means, with its L3 row also split by seat. For notebook composition, label proportions pool line counts across notebooks and divide by all original lines, dropped lines included. Annotation agreement is the fraction of lines on which the two initial labels agree, before adjudication. All sums, ratios, contrasts and intervals use unrounded values, and rounding is applied only for display.

\section{Repetition-level results}\label{sec:s-results}
This appendix follows the order of Section~4 of the paper. All summaries use the repetition as the unit, as in Appendix~\ref{sec:s-stats}.

\subsection{Breadth screen by seat (Section~4.2)}\label{sec:s-res-screen}
On L3 the breadth screen replaced the initiator in five repetitions and the responder in the other five. Table~\ref{tab:s-screen} splits its L3 row by the replaced seat, and the two groups average to the branch means of Table~1 of the paper.

\begin{table}[t]
\caption{Breadth screen on L3, split by the replaced seat: branch means over the five repetitions that replaced each seat, and the gain E $-$ F with its 95\% interval and the number of repetitions with the sign of its mean. The pooled row is the L3 row of Table~1 of the paper.}
\label{tab:s-screen}
\small
\setlength{\tabcolsep}{2.5pt}
\begin{tabular*}{\columnwidth}{@{\extracolsep{\fill}}lrrrrrcr@{}}
\toprule
Seat & Orig. & Exp. & Fresh & Loss & Gain & 95\% interval & $k$ \\
\midrule
Initiator & 70.90 & 60.80 & 63.50 & 10.10 & $-2.70$ & $[-8.32, 2.92]$ & 4/5 \\
Responder & 71.46 & 67.44 & 52.42 & 4.02 & 15.02 & $[8.61, 21.43]$ & 5/5 \\
\midrule
Pooled & 71.18 & 64.12 & 57.96 & 7.05 & 6.16 & & \\
\bottomrule
\end{tabular*}
\end{table}

\subsection{Formation length and changed agents (Section~4.3)}\label{sec:s-res-formation}
Table~\ref{tab:s-config} gives the continuity loss of each non-planning row of the configuration study with its interval.

\begin{table}[t]
\caption{Continuity loss O $-$ E in the configuration study on L3 (Table~2 of the paper): mean over ten repetitions and 95\% interval. The planning rows are compared through their paired reduction in Table~\ref{tab:s-summary}.}
\label{tab:s-config}
\setlength{\tabcolsep}{3.5pt}
\begin{tabular*}{\columnwidth}{@{\extracolsep{\fill}}llrc@{}}
\toprule
Factor & Setting & Loss & 95\% interval \\
\midrule
Formation length & 0 episodes & 0.02 & $[-2.57, 2.61]$ \\
 & 1 episode & 2.61 & $[0.48, 4.74]$ \\
 & 5 episodes (default) & 12.49 & $[-0.42, 25.39]$ \\
 & 10 episodes & 10.41 & $[8.76, 12.06]$ \\
\midrule
Model family & Gemini $\rightarrow$ Gemini & 9.19 & $[7.31, 11.07]$ \\
 & DeepSeek $\rightarrow$ DeepSeek & 10.87 & $[9.10, 12.65]$ \\
 & GPT $\rightarrow$ Gemini & 6.56 & $[2.85, 10.27]$ \\
 & DeepSeek $\rightarrow$ Gemini & 8.60 & $[6.11, 11.10]$ \\
\midrule
Temperature & $T = 0.0$ & 8.00 & $[4.20, 11.80]$ \\
 & $T = 1.0$ & 7.99 & $[6.32, 9.66]$ \\
\bottomrule
\end{tabular*}
\end{table}

\paragraph{Zero formation} With no formation episodes the donor notebook and the empty notebook are the same object, so the Experienced and Fresh branches are identical conditions. Their difference, $+0.88$ (SD 3.36, interval $[-1.53, 3.29]$, five repetitions on each side of zero), is therefore run-to-run variation alone. The loss in the same row is $+0.02$, interval $[-2.57, 2.61]$.

\paragraph{Five-episode row} This row reuses the seat study's ten L3 snapshots and Original evaluation and keeps the seat assigned to each repetition, five per seat, so that it is balanced like the other rows. Its loss is 14.45 where the initiator is replaced and 10.53 where the responder is, and it is positive in eight of the ten repetitions. Three repetitions lie far from the rest, repetitions 2, 3 and 7 with losses of $-14.94$, $40.76$ and $-15.12$, and they account for most of the row's spread and its wide interval, $[-0.42, 25.39]$. Without them the loss is 16.31, interval $[7.76, 24.85]$, positive in all seven remaining repetitions, in line with every other row of Table~\ref{tab:s-config} that includes formation. The conclusions of the seat study do not depend on these repetitions either. Without them, $\Dseat$ is 13.90, interval $[7.95, 19.85]$, for the whole notebook and 13.85, interval $[8.53, 19.17]$, for the partner half, both positive in all seven, and the rewording costs average 0.72 and 0.07 points in the initiator and responder seats. All other summaries keep every repetition.

\paragraph{Cross-model gains} The same Gemini newcomers, carrying the same donor notebooks, have a pooled gain of $+1.85$, interval $[-1.08, 4.78]$, in GPT teams and $-3.08$, interval $[-4.46, -1.70]$, in DeepSeek teams, where the gain is negative in all ten repetitions.

\subsection{Seat study and replication (Section~4.4)}\label{sec:s-res-seat}
Table~\ref{tab:s-summary} summarizes the contrasts of the seat study and of the content and planning studies over the ten repetitions. Table~\ref{tab:s-reps} lists every repetition of the seat asymmetries and rewording costs of Sections~4.4 and~4.5 of the paper, and Table~\ref{tab:s-second} reports the replication with twenty further pairs.

\begin{table*}[t]
\caption{Repetition-level summaries of the contrasts of the seat study and the content and planning studies: mean of the ten repetition-level values, their standard deviation (SD), the 95\% interval on the mean and the number of repetitions with the sign of the mean. Contrasts are in points of PC unless marked.}
\label{tab:s-summary}
\begin{tabular*}{\textwidth}{@{\extracolsep{\fill}}lrrcc@{}}
\toprule
Contrast & Mean & SD & 95\% interval & $k$/10 \\
\midrule
\multicolumn{5}{@{}l}{\textit{Seat asymmetry and its split, L3}} \\
Seat asymmetry $\Dseat$, whole notebook & 15.03 & 6.45 & $[10.41, 19.65]$ & 10/10 \\
Seat asymmetry $\Dseat$, partner half only & 15.10 & 6.03 & $[10.79, 19.42]$ & 10/10 \\
Seat asymmetry $\Dseat$, task half only & $-0.59$ & 2.62 & $[-2.47, 1.28]$ & 5/10 \\
Seat asymmetry $\Dseat$, partner half $-$ whole notebook & 0.07 & 2.13 & $[-1.45, 1.60]$ & 5/10 \\
Experienced term: responder branch $-$ initiator branch & 10.02 & 5.53 & $[6.06, 13.98]$ & 9/10 \\
\addlinespace[3pt]
\multicolumn{5}{@{}l}{\textit{Each half of the donor notebook against the whole, L3}} \\
Initiator: gain of the whole notebook & $-7.98$ & 4.84 & $[-11.44, -4.51]$ & 9/10 \\
Initiator: gain of the partner half & $-7.91$ & 5.00 & $[-11.49, -4.33]$ & 10/10 \\
Initiator: gain of the task half & 1.47 & 1.26 & $[0.57, 2.37]$ & 8/10 \\
Initiator: task half $-$ whole notebook & 9.45 & 5.22 & $[5.71, 13.19]$ & 10/10 \\
Initiator: partner half $-$ whole notebook & 0.07 & 1.70 & $[-1.15, 1.29]$ & 6/10 \\
Responder: gain of the whole notebook & 7.05 & 6.78 & $[2.20, 11.90]$ & 8/10 \\
Responder: gain of the partner half & 7.19 & 6.64 & $[2.44, 11.95]$ & 8/10 \\
Responder: gain of the task half & 0.88 & 1.65 & $[-0.30, 2.06]$ & 9/10 \\
Responder: task half $-$ whole notebook & $-6.17$ & 6.31 & $[-10.69, -1.66]$ & 8/10 \\
Responder: partner half $-$ whole notebook & 0.14 & 1.46 & $[-0.90, 1.19]$ & 7/10 \\
\addlinespace[3pt]
\multicolumn{5}{@{}l}{\textit{Paraphrase control, L3 (Para: the incumbent keeps its reworded notebook)}} \\
Initiator: cost of rewording, O $-$ Para & 0.52 & 0.85 & $[-0.08, 1.13]$ & 8/10 \\
Responder: cost of rewording, O $-$ Para & $-0.05$ & 0.87 & $[-0.67, 0.57]$ & 4/10 \\
\addlinespace[3pt]
\multicolumn{5}{@{}l}{\textit{Experience gain E $-$ F by level and replaced seat, default configuration}} \\
L1 initiator & 2.10 & 8.76 & $[-4.16, 8.36]$ & 8/10 \\
L1 responder & 2.61 & 10.44 & $[-4.86, 10.07]$ & 8/10 \\
L1 seat asymmetry & 0.50 & 5.43 & $[-3.38, 4.39]$ & 6/10 \\
L2 initiator & 4.44 & 4.11 & $[1.50, 7.38]$ & 9/10 \\
L2 responder & 6.48 & 6.63 & $[1.74, 11.22]$ & 10/10 \\
L2 seat asymmetry & 2.04 & 6.79 & $[-2.82, 6.90]$ & 5/10 \\
L3 initiator & $-7.98$ & 4.84 & $[-11.44, -4.51]$ & 9/10 \\
L3 responder & 7.05 & 6.78 & $[2.20, 11.90]$ & 8/10 \\
L3 seat asymmetry & 15.03 & 6.45 & $[10.41, 19.65]$ & 10/10 \\
\addlinespace[3pt]
\multicolumn{5}{@{}l}{\textit{Widening planning from two members to five, L3}} \\
Experienced, messages per unit of progress (\% change) & 13.33 & 3.71 & $[10.67, 15.98]$ & 10/10 \\
Experienced, tokens per unit of progress (\% change) & $-21.68$ & 31.37 & $[-44.12, 0.76]$ & 9/10 \\
Loss with two members $-$ loss with five & 6.40 & 3.19 & $[4.11, 8.68]$ & 10/10 \\
\bottomrule
\end{tabular*}
\end{table*}

\begin{table*}[t]
\caption{Every repetition of the seat asymmetries and rewording costs of Sections~4.4 and~4.5 of the paper, in points.}
\label{tab:s-reps}
\begin{tabular*}{\textwidth}{@{\extracolsep{\fill}}lrrrrrrrrrr@{}}
\toprule
Contrast & 1 & 2 & 3 & 4 & 5 & 6 & 7 & 8 & 9 & 10 \\
\midrule
$\Dseat$, whole notebook & 26.04 & 24.31 & 18.24 & 18.42 & 8.26 & 11.72 & 10.44 & 11.03 & 7.92 & 13.91 \\
$\Dseat$, partner half & 24.56 & 25.46 & 16.48 & 16.91 & 8.68 & 10.86 & 12.12 & 8.20 & 12.03 & 15.72 \\
$\Dseat$, task half & $-2.46$ & 1.28 & 2.76 & 0.87 & $-6.28$ & $-1.20$ & $-1.86$ & $-1.10$ & 1.92 & 0.16 \\
Rewording cost, initiator & 1.15 & 0.10 & $-0.02$ & 0.97 & 2.20 & 0.20 & 0.13 & 1.10 & $-0.88$ & 0.28 \\
Rewording cost, responder & $-0.29$ & $-0.38$ & 0.57 & $-1.65$ & 1.33 & 0.11 & $-1.16$ & 0.44 & 0.15 & 0.39 \\
\bottomrule
\end{tabular*}
\end{table*}

\begin{table}[t]
\caption{Replication of the L3 seat study: twenty further receiver--donor pairs in the default configuration, with both seats replayed from each snapshot. Means over the twenty pairs and 95\% intervals; the loss averaged over both seats is the mean over pairs of the two seats' losses, and $\Dseat$ is positive in 19 of the 20 pairs.}
\label{tab:s-second}
\begin{tabular*}{\columnwidth}{@{\extracolsep{\fill}}lrc@{}}
\toprule
Contrast & Mean & 95\% interval \\
\midrule
Continuity loss O $-$ E, initiator & 15.60 & $[12.42, 18.78]$ \\
Continuity loss O $-$ E, responder & 6.90 & $[4.61, 9.19]$ \\
Continuity loss, both seats averaged & 11.25 & $[8.87, 13.63]$ \\
Experience gain E $-$ F, initiator & $-7.10$ & $[-9.64, -4.56]$ \\
Experience gain E $-$ F, responder & 6.40 & $[3.71, 9.09]$ \\
Seat asymmetry $\Dseat$ & 13.50 & $[10.05, 16.95]$ \\
\bottomrule
\end{tabular*}
\end{table}

\paragraph{Splitting the seat asymmetry} On L3 the Experienced term of Appendix~\ref{sec:s-contrasts} is 10.02 points, interval $[6.06, 13.98]$, positive in nine of ten repetitions, two thirds of $\Dseat$. The Fresh term contributes the remaining third, the difference between the Fresh branches, 62.02 when the initiator is replaced and 57.00 when the responder is.

\paragraph{Messages per episode} In the L3 seat study the team sends 25.3\% more messages per episode than in the Original branch when the initiator is replaced and 16.9\% more when the responder is. With the communication premiums of 70.45\% and 28.19\% (Table~3 of the paper), progress per episode falls in both seats.

\subsection{Notebook content (Section~4.5)}\label{sec:s-res-content}
The content blocks of Table~\ref{tab:s-summary} compare each half with the whole notebook seat by seat, and Table~\ref{tab:s-reps} lists the per-repetition seat asymmetries of the whole notebook and of each half, together with the rewording costs. Appendix~\ref{sec:s-content} describes the paraphrase control and the line taxonomy, and Table~\ref{tab:s-split} gives the composition of the twenty donor notebooks.

\subsection{Planning width (Section~4.6)}\label{sec:s-res-planning}
The last block of Table~\ref{tab:s-summary} summarizes the paired changes from two planning members to five. The loss falls by 6.40 points, interval $[4.11, 8.68]$, in all ten repetitions. Inside the Experienced branch, messages per unit of progress rise by 13.33\%, interval $[10.67, 15.98]$, in all ten repetitions, while tokens per unit of progress change by $-21.68$\%, interval $[-44.12, 0.76]$, and fall in nine of ten. The two measures therefore move in opposite directions in nine of the ten repetitions.

\section{Notebook-content interventions}\label{sec:s-content}
\subsection{Paraphrase control}\label{sec:s-paraphrase}
The incumbent keeps its seat and its own notebook, rewritten by GPT-5.6 Terra with the prompt in Appendix~\ref{sec:s-instrument-prompts}, which asks it to keep every fact, instruction and reference, to change only wording and, within a section, sentence order, and to stay within the same length budget. Teammates, task list, model, temperature and budget are as in the Original branch, and no notice is given, so only the surface form of one notebook differs.

\subsection{Line taxonomy and labelling}\label{sec:s-taxonomy}
Each line of a donor notebook is one numbered note with one main point and is classified as
\begin{itemize}
\item \emph{partner-referring} if its main assertion concerns the other agent: what it will do, needs to be told or expects in return, including agreed handoffs and message conventions, whether or not this is specific to one partner, or
\item \emph{task-procedural} if it states task facts or procedures that would apply whoever the partner is, such as recipes, environment rules, timings and action orders.
\end{itemize}
A line touching both takes the label of its main assertion, and a line with neither kind of content is dropped. For example, one donor initiator's notebook holds the partner-referring line \emph{``For the pumpkin episode, re-staging slices at [2,2] and dish at [2,3] enabled successful handoff; partner handled ingredient preparation while I handled oven, pot, filling, and serving.''}

\begin{table}[t]
\caption{Composition of the twenty donor notebooks on L3: lines per notebook by label.}
\label{tab:s-split}
\setlength{\tabcolsep}{3.2pt}
\begin{tabular*}{\columnwidth}{@{\extracolsep{\fill}}lrrrrcrrrr@{}}
\toprule
& \multicolumn{4}{c}{Initiator} & & \multicolumn{4}{c}{Responder} \\
\cmidrule{2-5}\cmidrule{7-10}
Rep. & All & Task & Partner & Drop & & All & Task & Partner & Drop \\
\midrule
1 & 7 & 3 & 4 & 0 & & 7 & 4 & 3 & 0 \\
2 & 9 & 5 & 4 & 0 & & 7 & 4 & 3 & 0 \\
3 & 8 & 3 & 4 & 1 & & 5 & 3 & 2 & 0 \\
4 & 8 & 4 & 4 & 0 & & 10 & 5 & 4 & 1 \\
5 & 6 & 4 & 2 & 0 & & 7 & 3 & 4 & 0 \\
6 & 5 & 3 & 2 & 0 & & 9 & 5 & 4 & 0 \\
7 & 8 & 4 & 3 & 1 & & 6 & 3 & 3 & 0 \\
8 & 6 & 4 & 2 & 0 & & 7 & 4 & 3 & 0 \\
9 & 7 & 3 & 4 & 0 & & 6 & 4 & 2 & 0 \\
10 & 6 & 4 & 2 & 0 & & 7 & 3 & 4 & 0 \\
\midrule
Total & 70 & 37 & 31 & 2 & & 71 & 38 & 32 & 1 \\
\bottomrule
\end{tabular*}
\end{table}

The twenty donor notebooks of the L3 seat study (ten repetitions, two roles) contain 141 lines. Gemini 3.5 Flash-Lite and DeepSeek V4 Flash, neither of which wrote the notebooks, labelled every line independently with the prompt in Appendix~\ref{sec:s-instrument-prompts}. They agreed on 121 lines (85.8\%). GPT-5.6 Terra resolved the remaining 20 with the adjudication prompt, and three lines with neither kind of content were dropped. The taxonomy was fixed before any half was transferred. Across all twenty notebooks the labels split into 75 task-procedural, 63 partner-referring and 3 dropped lines. Task-procedural and partner-referring lines make up 52.9\% and 44.3\% of initiator notebooks and 53.5\% and 45.1\% of responder notebooks (Table~\ref{tab:s-split}). Each half is assembled from the original lines that carry its label, in their original order, with nothing summarized or added, and is installed on its own in the replaced seat, with everything else, including the newcomer's notice, as in the Experienced branch.

\section{Prompts}\label{sec:s-prompts}
Text in double braces is filled in by the program at run time, and long lines are wrapped for display, with continuation lines indented. Where a call has a system message and a user message, both are given. Otherwise the prompt is followed by the template of the message that carries the call's input. In every agent prompt the notebook block holds the agent's notebook, which is empty for a fresh newcomer and at the start of formation, and the notice block holds the newcomer's line of Appendix~\ref{sec:s-settings} for a newcomer and nothing for any other member.

\subsection{Agents in the embodied settings}\label{sec:s-agent-prompts}
Every agent in Collab-Overcooked and CREW-Wildfire receives the same system message. Its role block is filled with the module for the agent's role, its rules and action format are taken from the benchmark, and the user message is filled at each decision.

\needspace{5\baselineskip}
\paragraph{System message}
\begin{Verbatim}
You are {{agent_id}}, the {{role_name}} in a
  cooperative team.

ROLE RESPONSIBILITIES
{{role_instructions}}

ENVIRONMENT RULES
{{benchmark_rules}}

ACTION FORMAT
{{action_schema}}

PRIVATE NOTEBOOK
{{notebook_or_empty}}

{{newcomer_notice_or_empty}}

Work toward the shared task objective within your
  role. Use your permitted observations, the public
  messages, and your notebook. The current
  observation and environment feedback describe the
  state of this episode.

At each decision, choose exactly one action from the
  supplied action interface, using its action names
  and argument formats. Your public message may
  request work, provide information, or report a
  relevant change; it may also be empty. Put your
  brief private justification in the rationale
  field. Only your message is shared with teammates.

Return only one JSON object with exactly these
  fields:
- action: one native action in the supplied
  ACTION FORMAT;
- message: a public message string, or an empty
  string;
- rationale: a brief private justification string.
Do not return a sequence of actions.
\end{Verbatim}

\needspace{5\baselineskip}
\paragraph{User message, filled at each decision}
\begin{Verbatim}
TASK INFORMATION AVAILABLE TO YOU
{{permitted_task_context}}

YOUR CURRENT OBSERVATION
{{own_observation}}

PUBLIC MESSAGES AVAILABLE AT THIS DECISION
{{shared_messages}}

LATEST ENVIRONMENT FEEDBACK
{{latest_feedback}}

AVAILABLE ACTIONS AND ARGUMENTS
{{legal_actions}}

Choose your next action and public message.
\end{Verbatim}

\needspace{5\baselineskip}
\paragraph{Role module, Collab-Overcooked initiator}
\begin{Verbatim}
You are the initiator. You alone receive the recipe
  for this episode. Coordinate preparation with the
  responder and operate the pot, oven, and serving
  actions available to your role.

Use the recipe, current observation, and environment
  feedback to choose your next action. Communicate
  the ingredient, preparation, dish, or handoff
  requirements the responder needs to carry out its
  role. Actions assigned to the responder are
  handled through the supplied collaboration
  interface. Do not treat a request as evidence that
  the work is complete.
\end{Verbatim}

\needspace{5\baselineskip}
\paragraph{Role module, Collab-Overcooked responder}
\begin{Verbatim}
You are the responder. Prepare ingredients and
  dishes and hand them over to the initiator using
  the actions available to your role. You do not
  receive the recipe directly.

Use the initiator's public messages, your
  observation, and environment feedback to determine
  what preparation is needed. Carry out feasible
  preparation and handoff work. Communicate
  readiness, blocked work, or a missing requirement
  when useful for the team's next decision.
\end{Verbatim}

\needspace{5\baselineskip}
\paragraph{Role module, CREW-Wildfire acting agent}
\begin{Verbatim}
You are one of three acting agents with identical
  capabilities in a cooperative tree-cutting task.
  All three agents work toward the same objective in
  the supplied target cells.

Use the observed map, target information, agent
  state, public messages, and environment feedback
  to choose your next action. Coordinate intended
  work, relevant obstacles, and completed work when
  useful. Choose only an action from the supplied
  environment interface.
\end{Verbatim}

\subsection{MedQA team}\label{sec:s-medqa-prompts}
A round has four steps: each specialist analyses the question, the coordinator synthesizes the three analyses, each specialist reviews the synthesis, and the coordinator decides. The team stops when all three reviewers agree or after three rounds. Only specialist seats are replaced, so only the specialist prompts carry the notice block.

\needspace{5\baselineskip}
\paragraph{Specialist analysis} System message:
\begin{Verbatim}
You are {{specialist_id}}, a specialist in
  {{specialty}} in a medical multiple-choice
  question-answering team.

PRIVATE NOTEBOOK
{{notebook_or_empty}}

{{newcomer_notice_or_empty}}

Evaluate the supplied question and options from your
  specialty. Use the notebook as experience from
  previous episodes and the supplied earlier
  discussion as evidence from this episode. Analyze
  the question and distinguish the answer options.
  State your preferred option and a concise
  explanation of the clinical evidence supporting
  it. Give your own analysis before receiving the
  other specialists' analyses from this phase.

Use only the supplied answer labels. Return only:
{"answer": "one supplied option label",
 "explanation": "public explanation"}
\end{Verbatim}
\needspace{5\baselineskip}
\noindent User message:
\begin{Verbatim}
QUESTION
{{question}}

ANSWER OPTIONS
{{answer_options}}

ROUND
{{round_number}}

YOUR EARLIER ANALYSIS AND PERMITTED DISCUSSION
{{own_earlier_analysis_and_permitted_
  discussion_or_empty}}

Give your analysis for this round.
\end{Verbatim}

\needspace{5\baselineskip}
\paragraph{Coordinator synthesis} System message:
\begin{Verbatim}
You are the coordinator of a medical
  question-answering team with three specialists.

PRIVATE NOTEBOOK
{{notebook_or_empty}}

Read the question, answer options, and the three
  supplied specialist analyses. Synthesize their
  analyses into a proposed answer and explanation.
  Identify any important disagreement that the
  specialists should examine when reviewing the
  synthesis. Use a supplied option label. Do not use
  or request the reference answer.

Return only:
{"answer": "one supplied option label",
 "explanation": "public synthesis"}
\end{Verbatim}
\needspace{5\baselineskip}
\noindent User message:
\begin{Verbatim}
QUESTION
{{question}}

ANSWER OPTIONS
{{answer_options}}

ROUND
{{round_number}}

THREE SPECIALIST ANALYSES FROM THIS ROUND
{{current_specialist_analyses}}

EARLIER ROUND DISCUSSION
{{earlier_round_discussion_or_empty}}

Synthesize the analyses into a proposed answer for
  review.
\end{Verbatim}

\needspace{5\baselineskip}
\paragraph{Specialist review} System message:
\begin{Verbatim}
You are {{specialist_id}}, a specialist in
  {{specialty}} in a medical multiple-choice
  question-answering team.

PRIVATE NOTEBOOK
{{notebook_or_empty}}

{{newcomer_notice_or_empty}}

Review the coordinator's synthesis using the
  question, answer options, and your analysis. State
  whether you agree with its conclusion. If you
  disagree, identify the consequential issue and
  give your supported alternative. Do not report
  agreement merely because other specialists might
  agree.

Use only the supplied answer labels. Return only one
  JSON object with these fields:
- agree: a JSON boolean, true if you agree with the
  synthesis's conclusion and false otherwise;
- answer: one supplied option label;
- reason: your public review explaining your
  assessment.
\end{Verbatim}
\needspace{5\baselineskip}
\noindent User message:
\begin{Verbatim}
QUESTION
{{question}}

ANSWER OPTIONS
{{answer_options}}

ROUND
{{round_number}}

YOUR ANALYSIS FROM THIS ROUND
{{own_current_analysis}}

COORDINATOR SYNTHESIS FROM THIS ROUND
{{current_synthesis}}

EARLIER PERMITTED DISCUSSION
{{earlier_permitted_discussion_or_empty}}

Review the synthesis and state whether you agree.
\end{Verbatim}

\needspace{5\baselineskip}
\paragraph{Coordinator decision} The controller status states whether all three reviewers agreed and whether another round remains. System message:
\begin{Verbatim}
You are the coordinator of a medical
  question-answering team with three specialists.

PRIVATE NOTEBOOK
{{notebook_or_empty}}

Assess the specialists' reviews and provide the team
  answer for this round. If the controller reports
  that all three reviewers agree, provide the final
  answer. If another round is available, identify
  the unresolved issue in your explanation. At the
  three-round limit, provide a final answer by
  weighing the majority-supported specialist
  position together with the clinical evidence in
  the question and discussion.

Use a supplied option label. Do not use or request
  the reference answer.

CONTROLLER STATUS
{{controller_status}}

Return only:
{"answer": "one supplied option label",
 "explanation": "public decision explanation"}
\end{Verbatim}
\needspace{5\baselineskip}
\noindent User message:
\begin{Verbatim}
QUESTION
{{question}}

ANSWER OPTIONS
{{answer_options}}

ROUND
{{round_number}}

COORDINATOR SYNTHESIS FROM THIS ROUND
{{current_synthesis}}

THREE SPECIALIST REVIEWS FROM THIS ROUND
{{current_specialist_reviews}}

EARLIER ROUND DISCUSSION
{{earlier_round_discussion_or_empty}}

Provide the team answer for this round.
\end{Verbatim}

\subsection{Shared planning interface}\label{sec:s-planning-prompts}
In the planning variants every member, acting or planning-only, contributes through this call in the planning phase, before the acting interface is called. A member's allocated budget is its share of the team's 128 planning tokens per round, and the member-specific block takes one of the two modules below.

\needspace{5\baselineskip}
\paragraph{System message}
\begin{Verbatim}
You are {{agent_id}}, the {{role_name}},
  participating in the team's shared planning phase.

PRIVATE NOTEBOOK
{{notebook_or_empty}}

{{newcomer_notice_or_empty}}

Use the permitted planning information and public
  messages to contribute a useful next coordination
  step, dependency, or clarification. Do not
  introduce information absent from your permitted
  input. Keep your contribution within
  {{allocated_planning_tokens}} tokens.

{{acting_or_planning_only_instruction}}

Return only your public planning message, or an
  empty string. This response does not execute an
  environment action.
\end{Verbatim}

\needspace{5\baselineskip}
\paragraph{User message}
\begin{Verbatim}
PERMITTED PLANNING INFORMATION
{{permitted_planning_input}}

PUBLIC MESSAGES
{{shared_messages}}

Provide your planning contribution.
\end{Verbatim}

\needspace{5\baselineskip}
\paragraph{Module for an acting member}
\begin{Verbatim}
You also act as the initiator or responder. Choose
  environment actions only when the acting interface
  is called after planning.
\end{Verbatim}

\needspace{5\baselineskip}
\paragraph{Module for a planning-only member}
\begin{Verbatim}
You are a planning-only member. The initiator and
  responder execute environment actions; you do not
  select or execute them.
\end{Verbatim}

\subsection{Notebook update during formation}\label{sec:s-notebook-prompt}
\needspace{5\baselineskip}
Used by every agent between formation episodes. It is not called during evaluation, when the notebooks are frozen. The budget is 1,000 tokens.
\begin{Verbatim}
Update your private notebook to help you perform
  your role and work with teammates in future
  episodes. This notebook will be your only memory
  carried across episodes; the full interaction
  history will not persist.

Use only the previous notebook, your own
  observations and actions, messages actually
  received, and the supplied outcome feedback.
  Decide what is worth keeping from this experience.
  This may include an effective procedure, a failure
  and its lesson, a dependency in your work, or an
  observed way of coordinating with a teammate. Keep
  only content you judge useful and supported.

When recording an interaction, preserve the relevant
  role, situation, and observed behavior.
  Distinguish what someone proposed, what they
  agreed to, and what actually happened. Describe an
  isolated observation as such rather than turning
  it into a universal rule. You need not record a
  convention if none was established. Do not invent
  a partner preference or infer agreement from
  silence.

Maintain a cumulative notebook rather than a
  transcript or a summary of only the latest
  episode. Keep earlier useful lessons, revise
  claims contradicted by new evidence, and preserve
  uncertainty where the cause of an outcome is
  unclear. Future episodes start from fresh task
  states: name the task when a lesson depends on it,
  and do not treat temporary object locations or
  completed work as persistent facts.

Write concise numbered notes in your own words, with
  one main point per note. Do not use a fixed number
  of lines, a fixed line length, or separate quotas
  for kinds of content. The entire notebook must fit
  within {{notebook_budget}} tokens, including
  numbering. Prioritize useful supported information
  within this one shared budget.

Return only the updated notebook. If there is
  nothing supported to retain, return an empty
  string.

Input:
YOUR ROLE AND TASK CONTEXT
{{role_and_task_context}}

PREVIOUS NOTEBOOK
{{previous_notebook_or_empty}}

YOUR EPISODE RECORD
{{own_episode_record}}

OUTCOME FEEDBACK
{{outcome_feedback}}
\end{Verbatim}

\subsection{Notebook-content interventions}\label{sec:s-instrument-prompts}
\needspace{5\baselineskip}
\paragraph{Paraphrase control} GPT-5.6 Terra rewrites the incumbent's own notebook, the one the Original branch uses.
\begin{Verbatim}
Rewrite the supplied notebook while preserving its
  content.

Keep every fact, instruction, reference,
  qualification, and uncertainty. Preserve role
  assignments, teammate references, object names,
  coordinates, numbers, timing, and the distinction
  between observed, agreed, suggested, and expected
  behavior. Do not strengthen a tentative
  observation into a rule or weaken an explicit
  agreement into a suggestion.

Change wording and, within a section, sentence
  order. Preserve section boundaries when present.
  Do not add, remove, merge, correct, generalize, or
  explain any information. Do not replace a
  statement about a particular teammate with a
  generic instruction.

Stay within the same total budget of
  {{notebook_budget}} tokens. Return only the
  rewritten notebook.

Input:
NOTEBOOK TO REWRITE
{{incumbent_notebook}}
\end{Verbatim}

\needspace{5\baselineskip}
\paragraph{Line labelling} The same prompt is given separately to Gemini 3.5 Flash-Lite and DeepSeek V4 Flash, and neither receives the other's labels or any experimental result.
\begin{Verbatim}
Classify each numbered notebook line by its content.
  Assign exactly one label:

TASK: task facts or procedures that would apply
  regardless of the particular teammate, such as
  recipes, environment rules, action requirements,
  processing times, or action order. Merely
  mentioning an assigned role does not by itself
  make a task rule partner-specific.

PARTNER: statements about a teammate's behavior,
  what the teammate will do, what they need to be
  told, what they expect, or an interaction
  convention such as an agreed handoff or message
  meaning. The statement need not be unique to that
  teammate; classify what it refers to, not whether
  it could also hold for another teammate.

DROP: a line with no substantive task or partner
  content, such as a heading or uninterpretable
  fragment.

Read surrounding lines for context, but label every
  line separately. If a line concerns both task and
  partner content, assign the label of its main
  assertion. Do not split, rewrite, duplicate, or
  merge lines. Label the original statement,
  including its uncertainty, rather than an inferred
  explanation for it. Do not infer labels from
  whether a statement seems helpful, harmful,
  correct, or likely to improve a team.

Return exactly one JSON object:
{"labels":[{"line_id":"the supplied
  identifier","label":"TASK or PARTNER or
  DROP","reason":"a short content-based
  justification"}]}
Include every supplied line exactly once.

Input:
NOTEBOOK ROLE
{{role}}

NUMBERED NOTEBOOK LINES
{{lines_with_stable_ids}}
\end{Verbatim}

\needspace{5\baselineskip}
\paragraph{Adjudication of disagreements} GPT-5.6 Terra resolves the lines on which the two labellers disagree. The labelling definitions it receives are the TASK, PARTNER and DROP definitions of the labelling prompt above.
\begin{Verbatim}
Resolve disagreements between two independent
  notebook annotations. Use the supplied labeling
  definitions. You are deciding what each line is
  about, not whether transferring it helps or harms
  performance.

Read the original line and its notebook context
  before considering the two proposed labels. Choose
  exactly one final label: TASK, PARTNER, or DROP.
  If both content types occur, classify the main
  assertion. Preserve the original line without
  splitting, rewriting, merging, or duplication.

Do not favor either annotator or target a particular
  label count or agreement rate. Return exactly:
{"decisions":[{"line_id":"the supplied
  identifier","label":"TASK or PARTNER or
  DROP","reason":"a short justification grounded in
  the original text"}]}

Input:
LABELING DEFINITIONS
{{task_partner_drop_definitions}}

ORIGINAL NOTEBOOK
{{numbered_notebook}}

DISAGREEMENTS
{{line_id_and_two_annotations}}
\end{Verbatim}

\end{document}